\documentclass[numbib]{imaiai}
\usepackage{graphicx}
\usepackage{subfig}
\usepackage{booktabs}
\usepackage{amsfonts,amsmath,latexsym,amssymb,mathrsfs,amsthm}

\usepackage[sort&compress]{natbib}
\received{}
\begin{document}
\title{Identifying networks with common organizational principles}
\author{
{\sc Anatol E. Wegner}$ˆ*$,\\[2pt]
University College London, Department of Statistical Science,
Gower Street,
London WC1E 6BT,
UK\\University of Oxford, Department of Statistics, 24-29 St. Giles', Oxford, OX1 3LB, UK\\
$ˆ*$\email{Corresponding author: a.wegner@ucl.ac.uk}\\[6pt]
{\sc Luis Ospina-Forero}\\[2pt]
University of Oxford, Department of Statistics, 24-29 St. Giles', Oxford, OX1 3LB, UK\\
{luis.ospinaforero@linacre.ox.ac.uk}\\[6pt]
{\sc Robert E. Gaunt}\\[2pt]
The University of Manchester, School of Mathematics, Manchester M13 9PL, UK\\
University of Oxford, Department of Statistics, 24-29 St. Giles', Oxford, OX1 3LB, UK\\
{robert.gaunt@manchester.ac.uk}\\[6pt]
{\sc Charlotte M. Deane}\\[2pt]
University of Oxford, Department of Statistics, 24-29 St. Giles', Oxford, OX1 3LB, UK\\
{deane@stats.ox.ac.uk}\\[6pt]
{\sc and}\\[6pt]
{\sc Gesine Reinert} \\[2pt]
University of Oxford, Department of Statistics, 24-29 St. Giles', Oxford, OX1 3LB, UK\\
{reinert@stats.ox.ac.uk}}
\maketitle

\begin{abstract}
{Many complex systems can be represented as networks, and the problem of network comparison is becoming increasingly relevant.  There are many techniques for network comparison, from simply comparing network summary statistics to sophisticated but computationally costly alignment-based approaches.  Yet it remains challenging to accurately cluster networks that are of a different size and density, but hypothesized to be structurally similar. In this paper, we address this problem by introducing a new network comparison methodology that is aimed at identifying common organizational principles in networks. The methodology is simple, intuitive and applicable in a wide variety of settings ranging from the functional classification of proteins to tracking the evolution of a world trade network.}
{networks $|$ network comparison $|$ machine learning $|$ earth mover's distance $|$ network topology}\\
\end{abstract}




\section{Introduction}

Many complex systems can be represented as networks, including friendships, the World Wide Web, global trade flows and protein-protein interactions \citep{newmanbook}. The study of networks has been a very active area of research in recent years, and in particular, network comparison has become increasingly relevant e.g$.$ \citep{wilson2008study,netal,2014waqar,2014yaveroglu}. Network comparison itself has many wide-ranging applications, for example, comparing protein-protein interaction networks could lead to increased understanding of underlying biological processes \citep{2014waqar}. Network comparison can also be used to study the evolution of networks over time and for identifying sudden changes and shocks.  

Network comparison methods have attracted increasing attention in the field of machine learning, where they are mostly referred to as graph kernels, and have numerous applications in personalized medicine e.g$.$ \citep{borgdisease}, computer vision and drug discovery e.g$.$ \citep{NCI}.  In the machine learning setting, the problem of interest is to obtain classifiers that can accurately predict the class membership of graphs.   

Methods for comparing networks range from comparison of summary statistics to sophisticated but computationally expensive alignment-based approaches \citep{migraal, netal,sana}. Real-world networks can be very large and are often inhomogeneous, which makes the problem of network comparison challenging, especially when networks differ significantly in terms of size and density. In this paper, we address this problem by introducing a new network comparison methodology that is aimed at comparing networks according to their common organizational principles.

The observation that the degree distribution of many real world networks is highly right skewed and in many cases approximately follows a power law has been very influential in the development of network science \citep{barabasi1999emergence}. Consequently, it has become widely accepted that the shape of the degree distribution (for example, binomial vs power law) is indicative of the generating mechanism underlying the network. In this paper, we formalize this idea by introducing a measure that captures the shape of distributions. The measure emerges from the requirement that a metric between forms of distributions should be invariant under rescalings and translations of the observables. Based on this measure, we then introduce a new network comparison methodology, which we call $NetEmd$.   

Although our methodology is applicable to almost any type of feature that can be associated to nodes or edges of a graph, we focus mainly on distributions of small connected subgraphs, also known as graphlets. Graphlets form the basis of many of the state of the art network comparison methods \citep{2007gdda,2014waqar,2014yaveroglu} and hence using graphlet based features allows for a comparative assessment of the presented methodology. Moreover, certain graphlets, called network motifs  \citep{2002milo}, occur much more frequently in many real world networks than is expected on the basis of pure chance. Network motifs are considered to be basic building blocks of networks that contribute to the function of the network by performing modular tasks and have therefore been conjectured to be favoured by natural selection. This is supported by the observation that network motifs are largely conserved within classes of networks \citep{milo2004superfamilies,wegner}.  

Our methodology provides an effective tool for comparing networks even when networks differ significantly in size and density, which is the case in most  applications. The methodology performs well on a wide variety of networks ranging from chemical compounds having as few as 10 nodes to tens of thousands of nodes in internet networks. The method achieves state of the art performance even when it is based on rather restricted sets of inputs that can be computed efficiently and hence scales favourably to networks with millions and even billions of nodes. The method also behaves well under network sub-sampling as described in \citet{SS}. The methodology further meets the needs of researchers from a variety of fields, from the social sciences to the biological and life sciences, by being computationally efficient and simple to implement.

We test the presented methodology in a large number of settings, starting with clustering synthetic and real world networks, where we find that the presented methodology outperforms state of the art graphlet-based network comparison methods in clustering networks of different sizes and densities. We then test the more fine grained properties of $NetEmd$ using data sets that represent evolving networks at different points in time. Finally, we test whether $NetEmd$ can predict functional categories of networks by exploring machine learning applications and find that classifiers based on $NetEmd$ outperform state-of-the art graph classifiers on several benchmark data sets. 
\section{A measure for comparing shapes of distributions}
Here we build on the idea that the information encapsulated in the shape of the degree distribution and other network properties reflects the topological organization of the network. From an abstract point of view we think of the shape of a distribution as a property that is invariant under linear deformations i.e$.$ translations and re-scalings of the axis. For example, a Gaussian distribution always has its characteristic bell curve shape regardless of its mean and standard deviation. Consequently, we postulate that any metric that aims to capture the similarity of shapes should be invariant under linear transformations of its inputs.  

Based on these ideas we define the following measure between distributions $p$ and $q$ that are supported on $\mathbb{R}$ and have non-zero, finite variances:
\begin{equation}\label{emdmet}
EMD^*(p,q)=\mathrm{inf}_{c\in\mathbb{R}}\left( EMD\big(\tilde{p}(\cdot+c),\tilde{q}(\cdot)\big)\right),
\end{equation}
where $EMD$ is the earth mover's distance and $\tilde{p}$ and $\tilde{q}$ are the distributions obtained by rescaling $p$ and $q$ to have variance 1. More precisely, $\tilde{p}$ is the distribution obtained from $p$ by the transformation $x\rightarrow \frac{x}{\sigma(p)}$, where $\sigma(p)$ is the standard deviation of $p$. Intuitively, $EMD$ (also known as the 1st Wasserstein metric \citep{emd1998} can be thought of as the minimal work, i.e$.$ mass times distance, needed to ``transport'' the mass of one distribution onto the other. For probability distributions $p$ and $q$ with support in $\mathbb{R}$ and bounded absolute first moment, the $EMD$ between $p$ and $q$ is given by $EMD(p,q)=\int_{-\infty}^\infty|F(x)-G(x)|\,\mathrm{d}x$, where $F$ and $G$ are the cumulative distribution functions of $p$ and $q$ respectively.

In principle, $EMD$ in Equation (\ref{emdmet}) can be replaced by almost any other probability metric $d$ to obtain a corresponding metric $d^*$. Here we choose $EMD$ because it is well suited to comparing shapes, as shown by its many applications in the area of pattern recognition and image retrieval \citep{emd1998}. Moreover, we found that $EMD$ produces superior results to classical $L^1$ and Kolmogorov distances, especially for highly irregular distributions that one frequently encounters in real world networks. 

\begin{figure}
 \centering
 \subfloat[]{\includegraphics[width=0.50\linewidth]{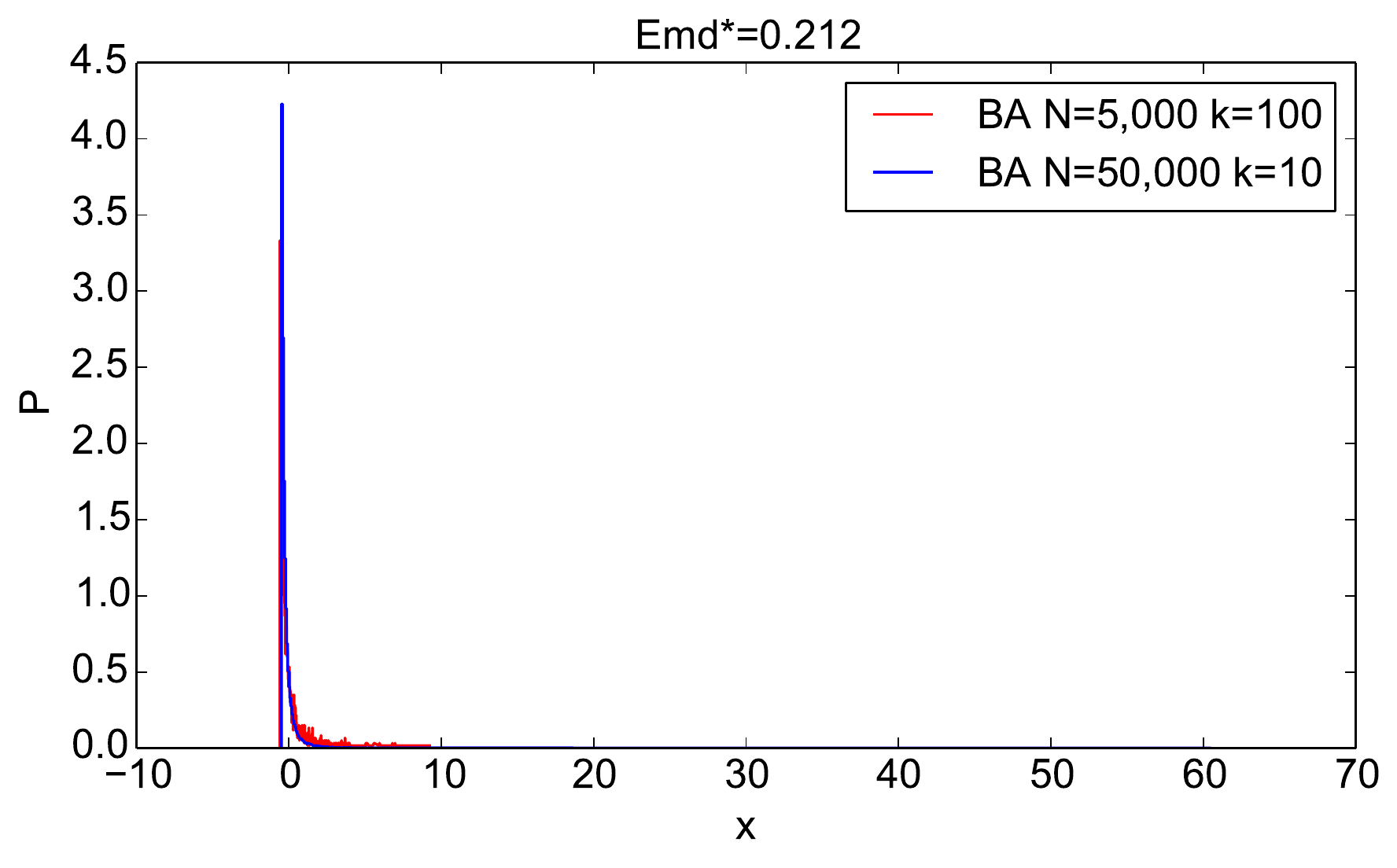}}
 \subfloat[]{\includegraphics[width=0.50\linewidth]{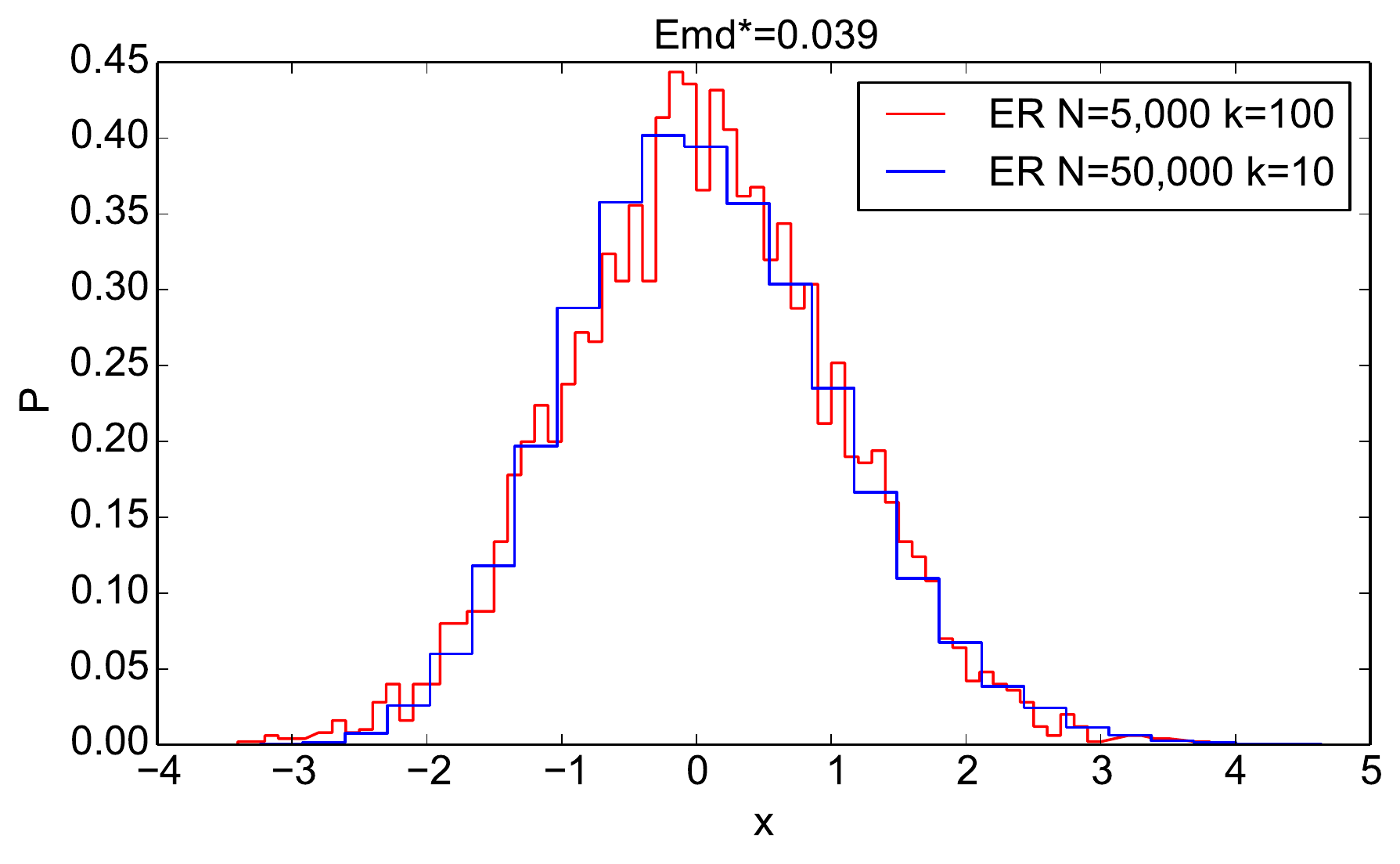}}
\quad
\subfloat[]
{\includegraphics[width=0.50\linewidth]{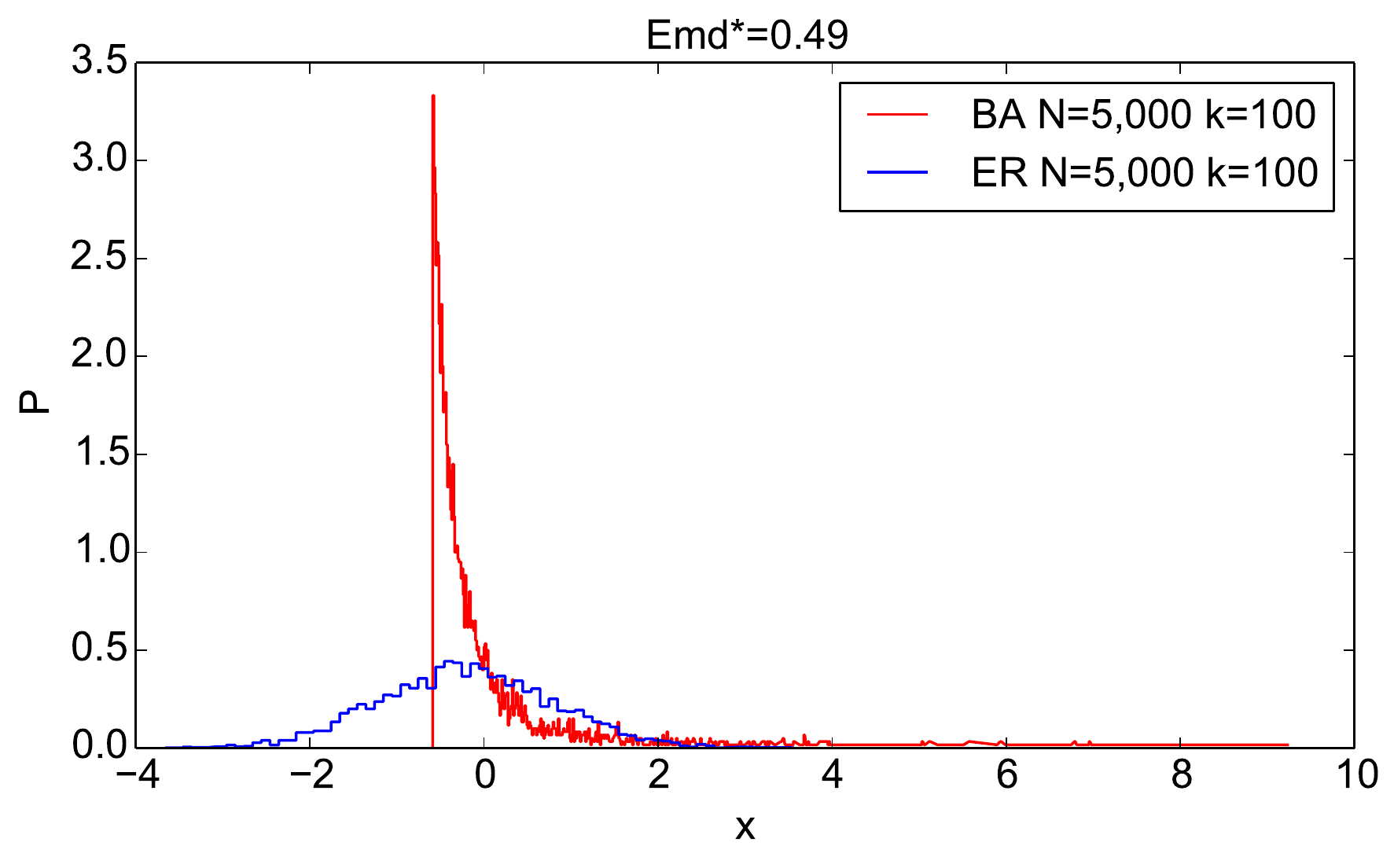}}
\caption{Plots of rescaled and translated degree distributions for Barabasi-Albert (BA) and Erd\H{o}s-R\'{e}nyi (ER) models with $N$ nodes and average degree $k$: a) BA $N=5,000$, $k=100$ vs BA $N=50,000$, $k=10$. b) ER $N=5,000$, $k=100$ vs ER $N=50,000$, $k=10$. c) BA $N=5,000$, $k=100$ vs ER $N=5,000$, $k=100$.  The $EMD^*$ distances between the degree distribution of two BA or ER models with quite different values of $N$ and $k$ are smaller than the $EMD^*$ distance between the degree distribution of a BA and ER model when the number of nodes and average degree are equal.}
\label{fig:DD}
\end{figure}

For two networks $G$ and $G'$ and given network feature $t$, we define the corresponding  $NetEmd_t$ measure by: 
\begin{equation}
NetEmd_t (G,G')=EMD^*(p_t(G),p_t(G')),
\end{equation}
where $p_t(G)$ and $p_t(G')$ are the distributions of $t$ on $G$ and $G'$ respectively. $NetEmd_t$ can be shown to be a pseudometric between graphs for any feature $t$ (see Sec$.$ \ref{metric}), that is it is non-negative, symmetric and satisfies the triangle inequality. Figure \ref{fig:DD} gives examples where $t$ is taken to be the degree distribution, and $p_t(G)$ is the degree distribution of $G$.  

Measures that are based on the comparison of multiple features  can be expected to be more effective at identifying structural differences between networks than measures that are based on a single feature $t$, because for two networks to be considered similar they must show similarity across multiple features. Hence, for a given set $T=\{t_1,t_2,...,t_m\}$ of network features, we define the $NetEmd$ measure corresponding to $T$ simply as: 
\begin{equation}\label{eq:def_netemd}
NetEmd_T(G,G')=\frac{1}{m}\sum_{j=1}^{m} NetEmd_{t_j} (G,G').
\end{equation}

Although $NetEmd$ can in principle be based on any set $T$ of network features to which one can associate distributions, we initially consider only features that are based on distributions of small connected subgraphs, also known as graphlets. Graphlets form the basis of many state of the art network comparison methods and hence allow for a comparative assessment of the proposed methodology.  

\begin{figure}[!h]
\centering
\includegraphics[width=0.85\textwidth]{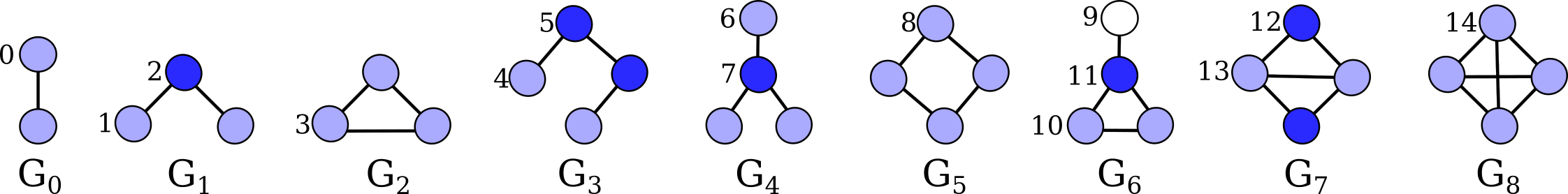}
\caption{Graphlets on two to four nodes. The different shades in each graphlet represent different automorphism orbits, numbered from 0 to 14.}
\label{fig.subgraphs_and_orbits}
\end{figure}

First, we consider graphlet degree distributions ($GDD$s) \citep{gdda} as our set of features. For a given graphlet $m$, the graphlet degree of a node is the number of graphlet-$m$ induced subgraphs that are attached to the node. One can distinguish between the different positions the node can have in $m$, which correspond to the automorphism orbits of $m$, see Figure \ref{fig.subgraphs_and_orbits}.  For graphlets up to size 5 there are 73 such orbits. We initially take the set of 73 $GDD$s corresponding to graphlets up to size 5 to be the default set of inputs, for which we denote the metric as $NetEmd_{G5}$. 

Later we also explore alternative definitions of subgraph distributions based on ego networks, as well as the effect of varying the size of subgraphs considered in the input. Finally, we consider the eigenvalue spectra of the graph Laplacian and the normalized graph Laplacian as inputs.  
\section{Results}
In order to give a comparative assessment of $NetEmd$, we consider to other graphlet based network comparison methods, namely $GDDA$ \citep{gdda}, $GCD$ \citep{2014yaveroglu} and Netdis \citep{2014waqar}. These represent the most effective alignment-free network comparison methodologies in the existing literature. While $GDDA$ directly compares distributions of graphlets up to size 5 in a pairwise fashion, $GCD$ is based on comparing rank correlations between graphlet degrees. Here we consider both default settings of GCD \citep{2014yaveroglu}, namely $GCD11$, which is based on a non-redundant subset of 11 graphlets up to size 4, and $GCD73$ which uses all graphlets up to size 5. $Netdis$ differs from $GDDA$ and $GCD$ in that it is based on subgraph counts in ego-networks of nodes. Another important distinction is that $Netdis$ first centers these raw counts by comparing them to the counts that could be expected under a particular null model before computing the final statistics. In our analysis, we consider two null models: an Erd\"os-R\'enyi random graph and a duplication divergence \citep{DD1} graph which has a scale-free degree distribution as well as a high clustering coefficient. We denote these two variants as $Netdis_{ER}$ and $Netdis_{SF}$ respectively. 
\subsection{Clustering synthetic and real world networks}
We start with the classical setting of network comparison where the task is to identify groups of structurally similar networks. The main challenge in this setting is to identify structurally similar networks even though they might differ substantially in terms of size and density. 
  
Given a set $S=\{G_1,G_2,...,G_n\}$ of networks consisting of disjoint classes $C=\{c_1,c_2,...,c_m\}$ one would like a network comparison measure $d$ to position networks from the same class closer to each other when compared to networks from other classes.  Given a network $G$, this can be measured in terms of the empirical probability $P(G)$ that $d(G,G_1)<d(G,G_2)$ where $G_1$ is a randomly selected network from the same class as $G$ (excluding itself) and $G_2$ is a randomly selected network from outside the class of $G$ and $d$ is the network comparison statistic. Consequently, the performance over the whole data set is measured in terms of the quantity $\overline{P}=\frac{1}{|S|}\sum_{G\in S}P(G)$.   It can be shown that $\overline{P}$ is equivalent to the average area under the receiver operator characteristic curve of a classifier that for a given network $G$ classifies the $k$ nearest neighbours of $G$ with respect to $d$ as being similar to $G$. Hence, a measure that positions networks randomly has an expected $\overline{P}$ of 0.5 whereas $\overline{P}=1$ corresponds to perfect separation between classes. Other measures are discussed in the Appendix. Conclusions reached in this paper hold regardless of which performance measure one uses.  

We first test $NetEmd$ on synthetic networks corresponding to realizations of eight random graph models, namely the Erd\H{o}s-R\'{e}nyi random graphs \citep{1960er}, the Barabasi Albert preferential attachment model \citep{barabasi1999emergence}, two duplication divergence models \citep{DD1,DD2}, the geometric gene duplication model \citep{2008higham}, 3D geometric random graphs \citep{2003penrose}, the configuration model \citep{1995molloy}, and Watts-Strogatz small world networks \citep{1998watts} (see Sec$.$ \ref{models} in the Appendix for details).

\begin{figure}[!htb]
 \centering
 \subfloat[Heatmap of $NetEmd_{G5}$ for $RG_2$.]{\includegraphics[width=0.45\linewidth]{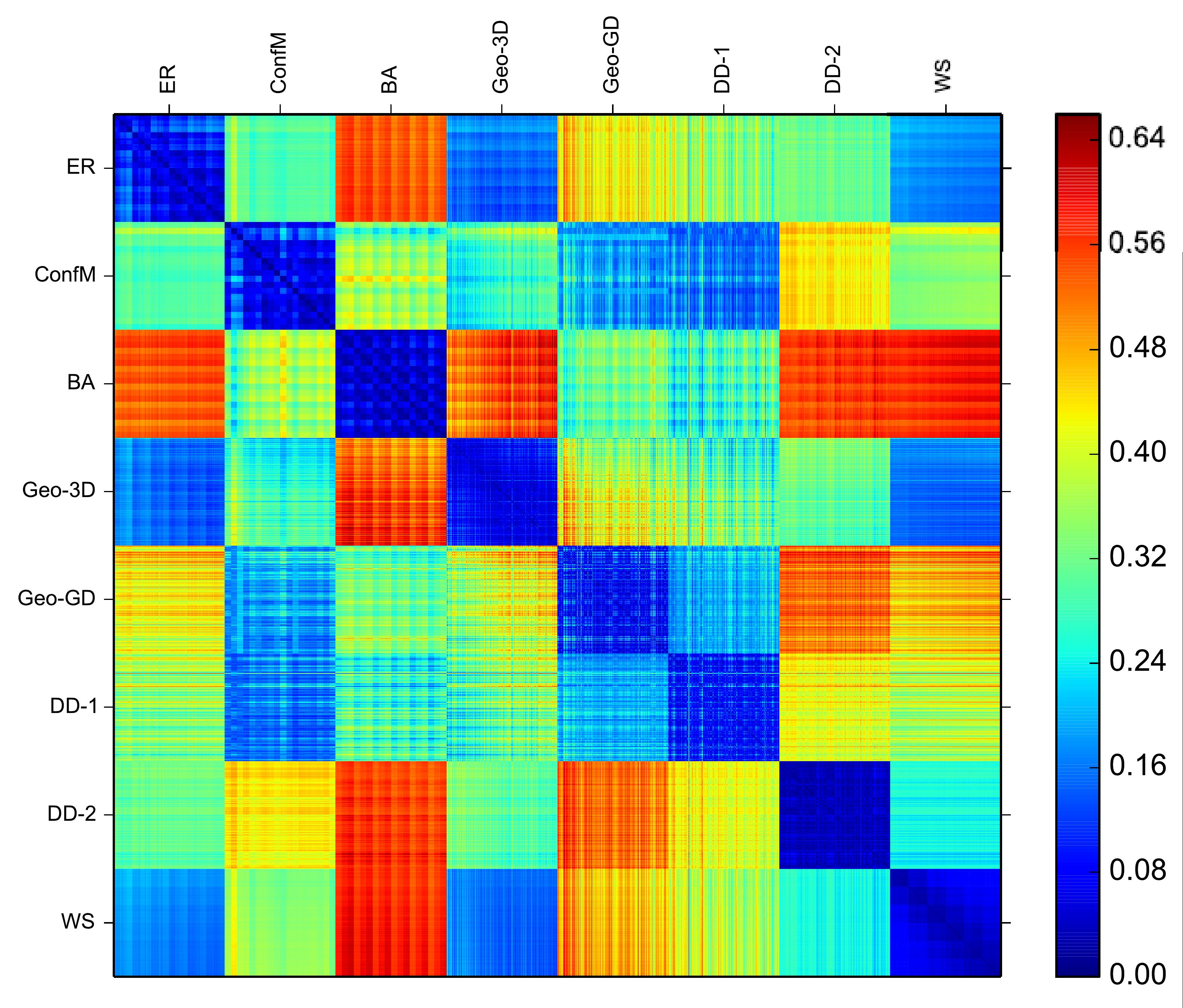}
 \label{rg2netemd}}\quad
\subfloat[Heatmap of $GCD73$ for $RG_2$.]{\includegraphics[width=0.44\linewidth]{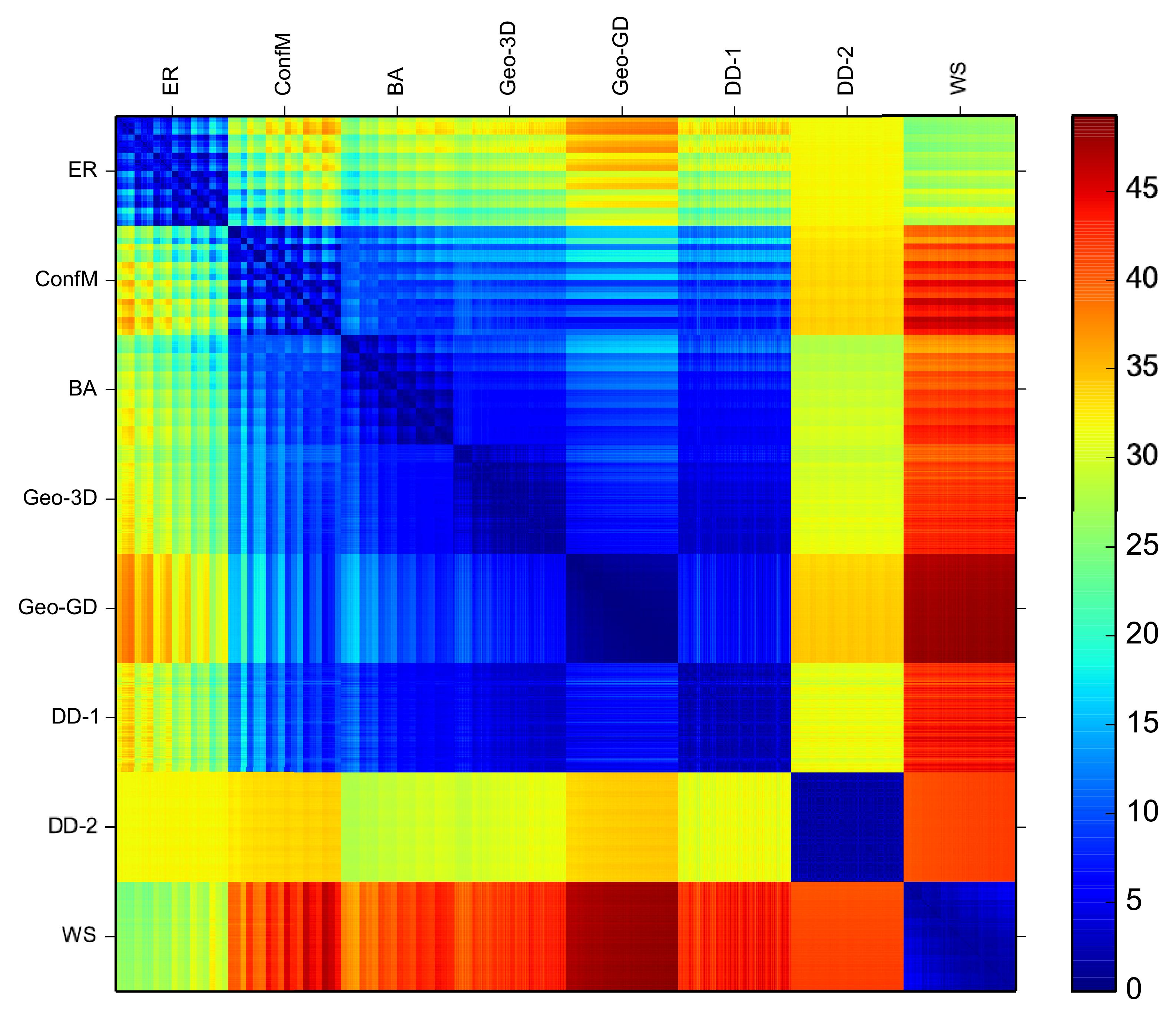}
\label{rg2gcd}}
\\
 \subfloat[$\overline{P}$ values for different network measures on data sets of synthetic and real world networks.]{\resizebox{\textwidth}{!}{{\begin{tabular}{lcccccc}
Dataset &$NetEmd_{G5}$&$Netdis_{ER}$&$Netdis_{SF}$ & $GCD11$ & $GCD73$ &  $GDDA$ \\
\midrule
Synthetic Networks&\\
$RG_1$ &\bf  0.997$\pm$0.003 \bf& 0.981$\pm$0.013 &0.986$\pm$0.011&0.992$\pm$0.012 &0.996$\pm$0.005&0.952$\pm$0.056\\
$RG_2$  &\bf 0.988\bf&0.897&0.919&0.976&0.976&0.956\\
$RG_3$&\bf0.925\bf&0.790&0.800&0.872&0.861 & 0.812\\
\midrule
RWN&\bf 0.942\bf& 0.898&0.866&0.898&0.906&0.745\\
\midrule
Onnela et al.&\bf0.890\bf&0.832&0.809&0.789& 0.819& 0.783\\
\bottomrule
\label{tab:clustering}
\end{tabular}}}
}
\caption{a) and b) show the heatmaps of pairwise distances on $RG_2$ ($N\in\{2000,3000,4000\}$ and $k\in\{20,24,28,32,36,40\}$) according to  $NetEmd_{G5}$ and $GCD73$, respectively. In the heat map, networks are ordered from top to bottom in the following order: model, average degree and node count. The heatmap of $NetEmd$ shows eight clearly identifiable blocks on the diagonal corresponding to different  generative models while the heatmap of $GCD73$ shows signs of off-diagonal mixing. c) $\overline{P}$ values for various comparison measures for data sets of synthetic and real world networks. For $RG_1$ we calculated the value of $\overline{P}$ for each of the 16 sub-data sets. The table shows the average and standard deviation of the $\overline{P}$ values obtained over these 16 sub-data sets. 
}
\label{fig:synthetic}
\end{figure}

For synthetic networks we consider three experimental settings of increasing difficulty, starting with the task of clustering networks that have same size $N$ and average degree $k$ according to generating mechanism - a task that is relevant in a model selection setting. For this we generate 16 data sets, which collectively we call $RG_1$, corresponding to combinations of $N\in\{1250,2500,5000,10000\}$ and $k\in\{10,20,40,80\}$, each containing 10 realizations per model, i.e. 80 networks. This is an easier problem than clustering networks of different sizes and densities, and in this setting we find that the $\overline{P}$ scores (see Table \ref{tab:clustering}) of top performing measures tend to be within one standard deviation of each other. We find that $NetEmd_{G5}$ and $GCD73$ achieve the highest scores, followed by $GCD11$ and $Netdis_{SF}$. 

Having established that $NetEmd$ is able to differentiate networks according to generating mechanism, we move on to the task of clustering networks of different sizes and densities. For this we generate two data sets: $RG_2$ in which the size $N$ and average degree $k$ are increased independently in linear steps to twice their initial value ($N\in\{2000,3000,4000\}$ and $k\in\{20,24,28,32,36,40\}$) and $RG_3$ in which the size and average degree are increased independently in multiples of 2 to 8 times their initial value ($N\in\{1250,2500,5000,10000\}$ and $k\in\{10,20,40,80\}$).  In $RG_3$, the number of nodes and average degrees of the networks both vary by one order of magnitude, and therefore clustering according to model type is challenging. Both $RG_2$ and $RG_3$ contain 10 realizations per model parameter i.e. contain $3\times6\times8\times10=1440$ and $4\times4\times8\times10=1280$ networks, respectively. Finally, we consider a data set consisting of networks from 10 different classes of real world networks (RWN) as well as a data set from \citep{2014waqar} that consists of real world and synthetic networks from the larger collection compiled by Onnela $et$ $al.$ \citep{onnela}.

We find that $NetEmd_{G5}$ outperforms all of the other three methods at clustering networks of different sizes and densities on all data sets. The difference can also be seen in the heatmaps of $NetEmd_{G5}$ and $GCD73$, the second best performing method for $RG_2$, given in Figures \ref{rg2netemd} and \ref{rg2gcd}. While the heatmap of $NetEmd_{G5}$ shows eight clearly identifiable blocks on the diagonal corresponding to different  generative models, the heatmap of $GCD73$ shows signs of off-diagonal mixing. The difference in performance becomes even more pronounced on more challenging data sets, i.e$.$ on $RG_3$ (see Fig$.$ \ref{fig:RG3} in the Appendix) and the Onnela $et$ $al.$ data set.

\subsection{Time ordered networks}
A network comparison measure should ideally not only be able to identify groups of similar networks but should also be able to capture structural similarity at a finer local scale. To study the behavior of $NetEmd$ at a more local level, we consider data sets that represent a system measured at different points in time. Since such networks can be assumed to evolve gradually over time they offer an ideal setting for testing the local properties of network comparison methodologies. 

We consider two data sets, named  AS-caida and AS-733 \citep{as}, that represent the topology of the internet at the level of autonomous systems and a third data set that consists of  bilateral trade flows between countries for the years 1962--2014 \citep{feenstra2005world,comtrade}. Both edges and nodes are added and deleted over time in all three data sets. As was noted in \citep{as} the time ranking in evolving networks is reflected to a certain degree in simple summary statistics. Hence, recovering the time ranking of evolving networks should be regarded as a test of consistency rather than an evaluation of performance. 

In order to minimize the dependence of our results on the algorithm that is used to rank networks, we consider four different ways of ranking networks based on their pairwise distances as follows. We assume that either the first or last network in the time series is given. Rankings are then constructed in a step-wise fashion. At each step one either adds the network that is closest to the last added network (Algorithm 1), or adds the network that has smallest average distance to all the networks in the ranking constructed so far (Algorithm 2). The performance of a measure in ranking networks is then measured in terms of Kendall's rank correlation coefficient $\tau$  between the true time ranking and the best ranking obtained by any of the 4 methods.
 \begin{figure}[!htb]
 \centering
 \subfloat[Heatmap of $NetEmd_{G5}$ for AS-733]{\includegraphics[width=0.275\linewidth]{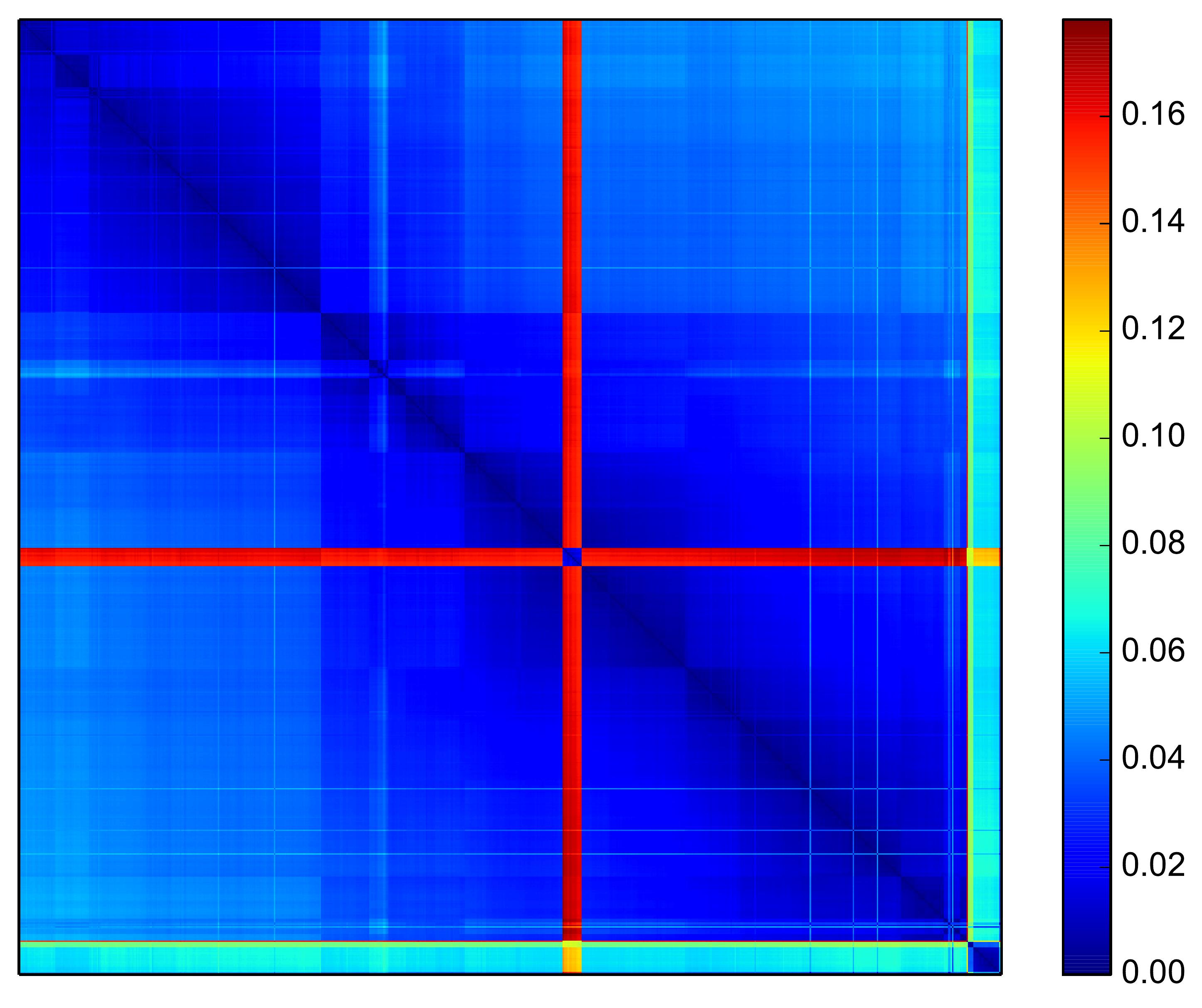}}
  \qquad
 \subfloat[Heatmap of $NetEmd_{G5}$ for AS-caida]{\includegraphics[width=0.275\linewidth]{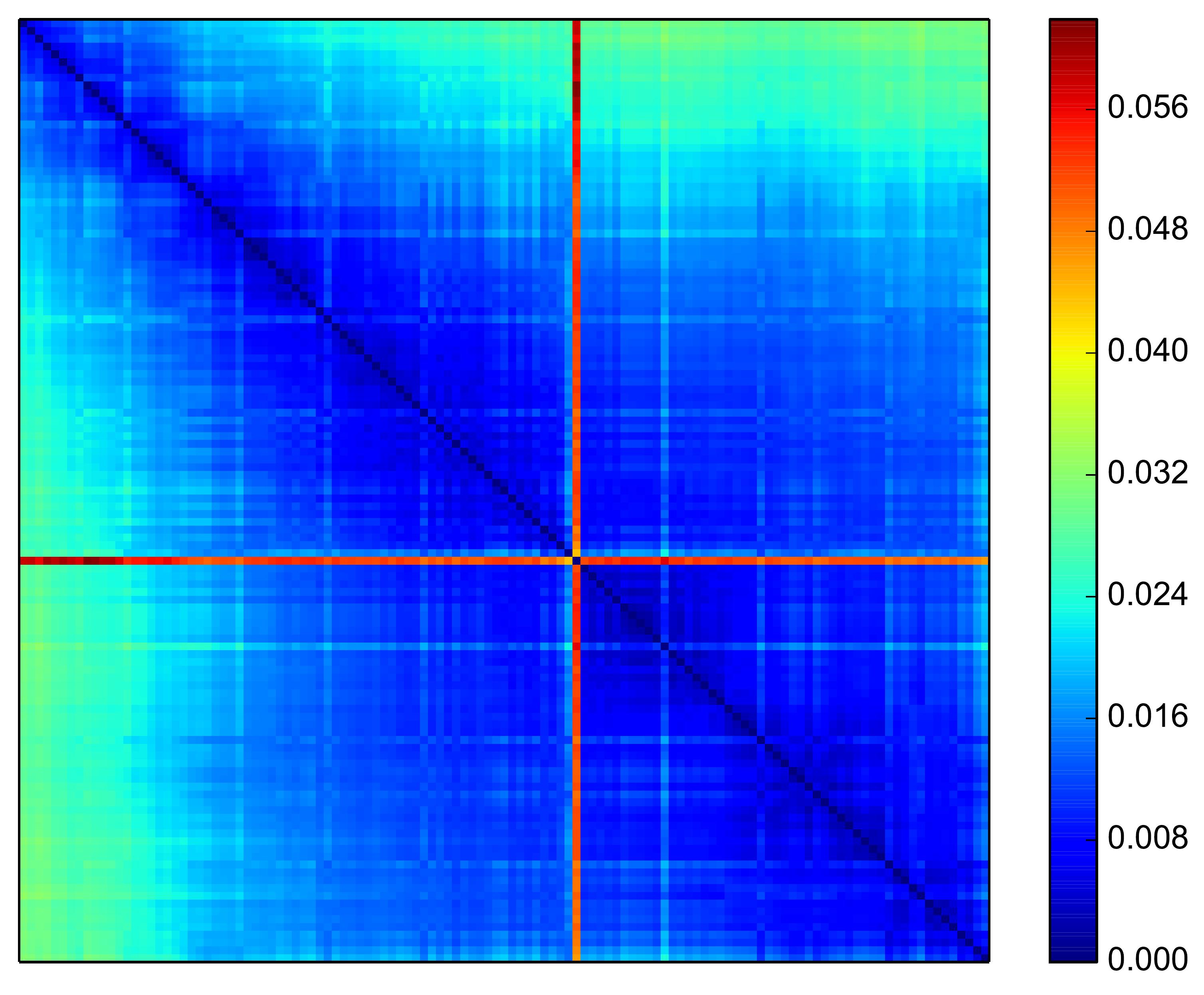}}
  \qquad
\subfloat[Heatmap of $NetEmd_{G5}$ for world trade networks.]{\includegraphics[width=0.30\linewidth]{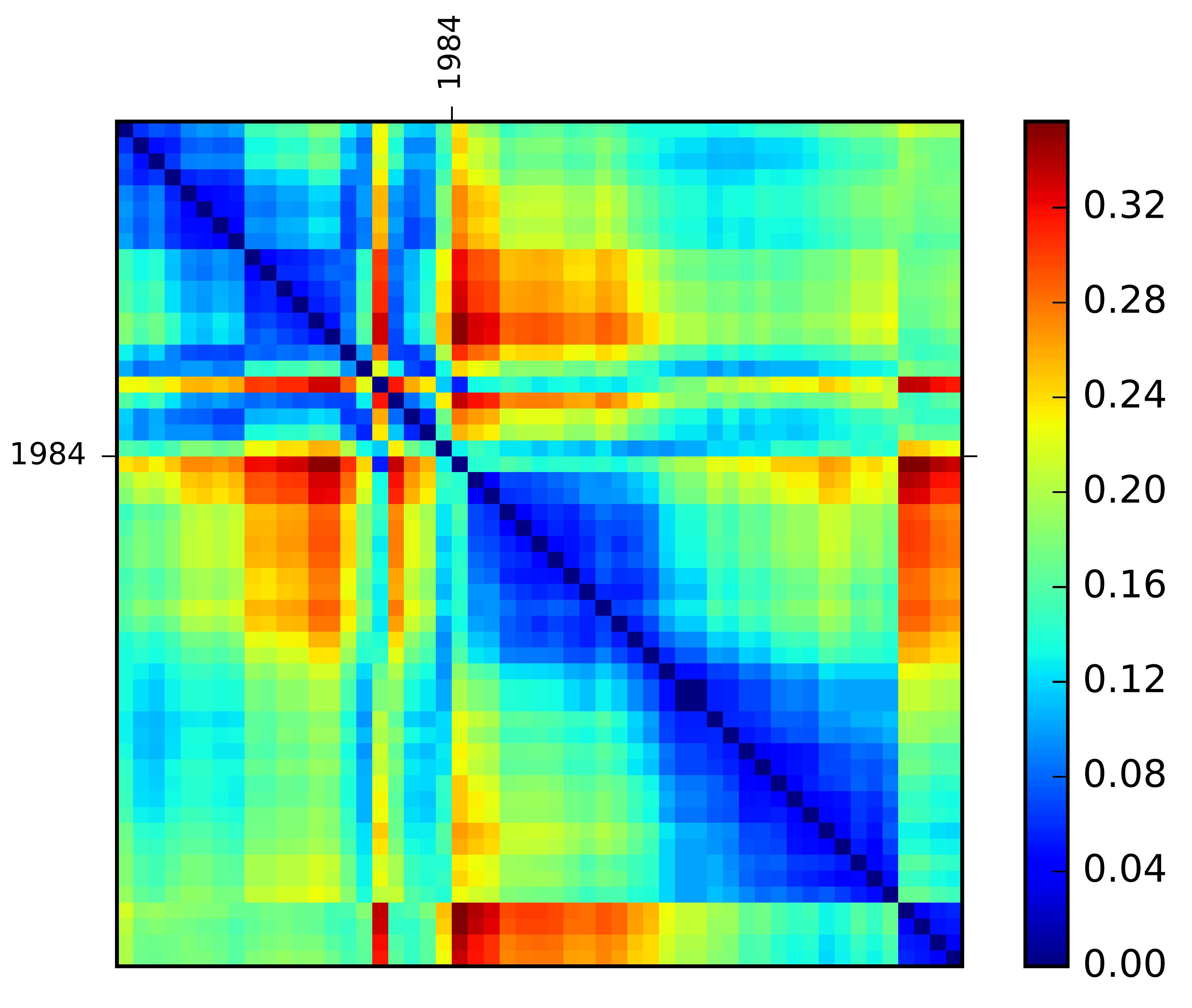}
\label{WTN}}\\
\subfloat[Kendall's $\tau$ between the true time ranking and rankings inferred from network comparison methodologies.]{\resizebox{\textwidth}{!}{{\begin{tabular}[b,width=0.3\linewidth]{lcccccc}
Dataset &$NetEmd_{G5}$&$Netdis_{ER}$&$Netdis_{SF}$ & $GCD11$ & $GCD73$ &  $GDDA$ \\
\midrule
AS-733&0.874&0.867&\bf0.933\bf&0.763&0.770&0.740\\
AS-caida &0.890& 0.844 & 0.849 &\bf0.897\bf&0.878 &0.870\\
World Trade &\bf0.821\bf&0.666&0.388& 0.380&0.567 &  0.649\\
\bottomrule

\end{tabular}}}

}
\caption{(a), (b) \& (c) Heatmaps of $NetEmd_{G5}$ for networks representing the internet at the level of autonomous systems networks and world trade networks. The date of measurement increases from left to right/ top to bottom. $NetEmd_{G5}$ accurately captures the evolution over time in all three data sets by positioning networks that are close in time closer to each other resulting in a clear signal along the diagonal.(d) Kendall's rank correlation coefficient between the true time ranking and rankings inferred from different network comparison measures.}
\label{time}

\end{figure}
We find that $NetEmd_{G5}$ successfully recovers the time ordering for all three data sets, as can be seen in the time ordered heatmaps given in Figure \ref{time} which all show clear groupings along the diagonal. The red regions in the two internet data sets correspond to outliers which can also be identified as sudden jumps in summary statistics e.g. the number of nodes. The two large clusters in the heatmap of world trade networks (Figure \ref{WTN}) coincide with a change in the data gathering methodology in 1984 \citep{feenstra2005world}. Although $NetEmd_{G5}$ comes second to $Netdis_{SF}$ on AS-733 and to $GCD11$ on AS-caida, $NetEmd_{G5}$ has the highest overall score and is the only measure that achieves consistently high scores on all three data sets.

\subsection{NetEmd based on different sets of inputs} We examine the effect of reducing the size of graphlets considered in the input of $NetEmd$, which is also relevant from a computational point of view, since enumerating graphlets up to size 5 can be challenging for very large networks. We consider variants based on the graphlet degree distributions of graphlets up to size 3 and 4, which we denote as $NetEmd_{G3}$ and $NetEmd_{G4}$. We also consider $NetEmd_{DD}$ which is based only on the degree distribution as a baseline. Results are given in Table \ref{tab:variants}. 

We find that reducing the size of graphlets from 5 to 4 does not significantly decrease the performance of $NetEmd$ and actually produces better results on three data sets ($RG_3$, Real world and Onnela et $al.$). Even when based on only graphlets up to size 3, i.e. just edges, 2-paths and triangles, $NetEmd$ outperforms all other non-$NetEmd$ methods that we tested on at least 6 out of 8 data sets.

Given that the complexity of enumerating graphlets up to size $s$ in a network on $N$ nodes having maximum degree $k_{mak}$  is $O(Nk_{max}^{s-1})$, $NetEmd_{G4}$ offers an optimal combination of performance and computational efficiency in most cases. The even less computationally costly $NetEmd_{G3}$ scales favourably even to networks of billions of edges for which enumerating graphlets of size 4 can be computationally prohibitive. This opens the door for comparing very large networks which are outside the reach of current methods while still retaining state of the art performance. Furthermore, the $NetEmd$ measures perform well under sub-sampling of nodes \citep{SS} (see Appendix D) which can be leveraged to further improve computational efficiency.  

We find that in some cases restricting the set of inputs actually leads to an increase in the performance of $NetEmd$. This indicates that not all graphlet distributions are equally informative in all settings \citep{2017_maugis}. Consequently, identifying (learning) which graphlet distributions contain the most pertinent information for a given task might lead to significant improvements in performance. Such generalizations can be incorporated into $NetEmd$ in a straightforward manner, for instance by modifying the sum in Equation (\ref{eq:def_netemd}) to incorporate weights. $NetEmd$ is ideally suited for such metric learning \citep{metriclearning} type generalizations since it constructs an individual distance for each graphlet distribution. Moreover, such single feature $NetEmd$ measures are in many cases highly informative even on their own. For instance $NetEmd_{DD}$, which only uses the degree distribution, outperforms the non-$NetEmd$ measures we tested individually on more than half the data sets we considered. 

We also considered counts of graphlets up to size 4 in 1-step ego networks of nodes ($NetEmd_{E4}$) \citep{2014waqar} as an alternative way of capturing subgraph distributions, for which we denote the measure as $NetEmd_{E4}$. Although we find that  $NetEmd_{E4}$ achieves consistently high scores, we find that variants based on graphlet degree distributions tend to perform better on most data sets. 

Finally, we consider spectral distributions of graphs as a possible  alternative to graphlet based features. The spectra of various graph operators are closely related to topological properties of graphs \citep{chung1997spectral,mohar1991laplacian,banerjee2008spectrum} and  have been widely used to characterize and compare graphs \citep{gu2016spectral,wilson2008study}. We used the spectra of the graph Laplacian and normalized graph Laplacian as inputs for $NetEmd$ for which we denote the measure as $NetEmd_S$. For a given graph the Laplacian is defined as $L=D-A$ where $A$ is the adjacency matrix of the graph and $D$ is the diagonal matrix whose diagonal entries are the node degrees. The normalized Laplacian $\hat{L}$ is defined as $D^{-\frac{1}{2}}LD^{-\frac{1}{2}}$. Given the eigenvalue distributions $S(L)$ and $S(\hat{L})$ of $L$ and $\hat{L}$ we define $NetEmd_S$ to be $\frac{1}{2}(NetEmd_{S(L)}+NetEmd_{S(\hat{L})})$.  

We find that in general $NetEmd_S$ performs better in clustering random graphs of different sizes and densities when compared to graphlet based network comparison measures. However, on the RWN and Onnela et al. data sets graphlet based $NetEmd$ measures tend to perform better than the spectral variant which can be attributed to the prevalence of network motifs in real world networks, giving graphlet based measures an advantage. The spectral variant is also outperformed on the time ordering of data sets which in turn might be a result of the sensitivity of graph spectra to small changes in the underlying graph \citep{wilson2008study}.

\begin{table}[!hbt]

{\resizebox{\textwidth}{!}{\begin{tabular}{lccccc}
Data set& $NetEmd_{G3}$ & $NetEmd_{G4}$ & $NetEmd_{E4}$&$NetEmd_{S}$&$NetEmd_{DD}$\\
\midrule

$RG_1$&0.989$\pm$0.008&0.995$\pm$0.005&0.993$\pm$0.004 &0.992$\pm$0.007&0.957$\pm$0.024\\

$RG_2$&0.982&0.987&0.983&\bf0.992\bf&0.944\\

$RG_3$&0.940&0.941& 0.947 &\bf0.972\bf&0.902\\
\midrule
RWN&\bf0.952\bf &0.950& 0.933&0.933&0.907\\

Onnela et al.&0.892& \bf0.898\bf &0.892&0.858&0.867\\
\midrule
AS-733&0.808&0.874& 0.922&0.855&0.928\\
AS-caida& \bf0.898\bf &0.892&0.820&0.780&0.821\\
World Trade &0.697& 0.785&0.665&0.430&0.358\\
\bottomrule
\end{tabular}}}
\caption{Results for different variants of $NetEmd$ based on distributions of graphlets up to size 3 and 4 ($NetEmd_{G3}$ and $NetEmd_{G4}$),  counts of graphlets up to size 4 in 1-step ego networks of nodes ($NetEmd_{E4}$), eigenvalue spectra of Laplacian operators ($NetEmd_{s}$) and the degree distribution ($NetEmd_{DD}$). Values in bold indicate that a measure achieves the highest score among all measures considered in the manuscript. For $RG_1$ we calculate the value of $\overline{P}$ for each of the 16 sub-data sets. The table shows the average and standard deviation of the $\overline{P}$ values obtained over these 16 sub-data sets. }
\label{tab:variants}
\end{table}
\subsection{Functional classification of networks} One of the primary motivations in studying the structure of networks is to identify topological features that can be related to the function of a network. In the context of network comparison this translates into the problem of finding metrics that can identify functionally similar networks based on their topological structure.

In order to test whether $NetEmd$ can be used to identify functionally similar networks, we use several benchmarks from the machine learning literature where graph similarity measures, called graph kernels, have been intensively studied over the past decade. In the context of machine learning the goal is to construct classifiers that can accurately predict the class membership of unknown graphs. 

We test $NetEmd$ on benchmark data sets representing social networks \citep{yanardag2015deep} consisting of Reddit posts, scientific collaborations and ego networks in the Internet Movie Database (IMDB). The Reddit data sets Reddit-Binary, Reddit-Multi-5k and Reddit-Multi-12k consist of networks representing Reddit treads where nodes correspond to users and two users are connected whenever one responded to the other's comments. While for the Reddit-Binary data sets the task is to classify networks into discussion based and question/answer based communities, in the data sets Reddit-Multi-5k and Reddit-Multi-12k the task is to classify networks according to their subreddit categories. COLLAB is  a  data set consisting of ego-networks of scientists from the fields High Energy Physics, Condensed Matter  Physics and Astro  Physics and the task is to determine which of these fields a given researcher belongs to. Similarly, the data sets IMDB-Binary and IMDB-Multi represent collaborations between film actors derived from the IMDB and the task is to classify ego-networks into different genres i.e. action and romance in the case of IMDB-Binary and comedy, action and Sci-Fi genres in the case of IMDB-Multi.   

We use C - support vector machine (C-SVM) \citep{csvm} classifiers with a Gaussian kernel $K(G,G')=\exp (-\alpha NetEmd(G,G')^2)$, where $\alpha$ is a free parameter to be learned during training. Performance evaluation is carried out by 10 fold cross validation, where at each step of the validation 9 folds are used for training and 1 fold for evaluation. Free parameters of classifiers are learned via 10 fold cross validation on the training data only. Finally, every experiment is repeated 10 fold and average prediction accuracy and standard deviation are reported. 

\begin{table}[ht]
\resizebox{\textwidth}{!}{\begin{tabular}[width=\linewidth]{lcccccc}
Kernel&Reddit-Binary& Reddit-Multi-5k& Reddit-Multi-12k & COLLAB& IMDB-Binary&IMDB-Multi\\
\midrule

$NetEmd_{G5}$ &\bf92.67 $\pm$0.30\bf&\bf54.61$\pm$0.18\bf &\bf48.09$\pm$0.21\bf&\bf79.32$\pm$0.27\bf&66.99 $\pm$1.19&41.45$\pm$0.70 \\
$NetEmd_S$ &88.59 $\pm$0.35&53.05$\pm$0.34&44.45$\pm$0.18&\bf79.05$\pm$0.20\bf&\bf71.68$\pm$0.88\bf&\bf46.06$\pm$0.50\bf \\
\midrule

DGK &78.04$\pm$0.39&41.27$\pm$0.18&32.22$\pm$0.10&73.09
$\pm$0.25&66.96
$\pm$0.56&44.55
$\pm$0.52\\
GK &77.34$\pm$0.18&41.01$\pm$0.17&31.82$\pm$0.08&72.84
$\pm$0.28&65.87
$\pm$0.98&43.89
$\pm$0.38\\

\midrule
RF&88.7$\pm$
1.99&50.9 $\pm$2.07&42.7 $\pm$1.28&76.5$\pm$1.68&\bf72.4 $\pm$4.68\bf&\bf47.8
$\pm$3.55\bf\\
\midrule
PCSN&86.30$\pm$1.58&49.10$\pm$0.70&41.32$\pm$0.42&72.60$\pm$2.15&\bf71.00$\pm$2.29\bf&\bf45.23$\pm$2.84\\

\bottomrule
\end{tabular}}
\caption{10 fold cross validation accuracies of Gaussian kernels based on $NetEmd$ measures using the distributions of graphlets up to size 5 ($NetEmd_{G5}$) and Laplacian spectra ($NetEmd_S$) and other graph kernels, namely the  deep graphlet kernels (DGK)\citep{yanardag2015deep} and the graphlet kernel (GK) \citep{GK}. We also consider alternatives to support vector machines classifiers, namely the random forest classifiers (RF) introduced in \citep{featurebased} and convolutional neural networks (PCSN) \citep{PCSN}. Values in bold correspond to significantly higher scores, which are scores with t-test p-values less than 0.05 when compared to the highest score.}
\label{usvm}
\end{table}

Table \ref{usvm} gives classification accuracies obtained using $NetEmd$ measures based on graphlets up to size five ($NetEmd_{G5}$) and spectra of Laplacian operators ($NetEmd_{S}$) on the data sets representing social networks. We compare $NetEmd$ based kernels to  graphlet kernels \citep{GK} and deep graphlet kernels \citep{yanardag2015deep} as well as two non-SVM classifiers namely the random forest classifier introduced in \citep{featurebased} and the  convolutional neural network based classifier introduced in \citep{PCSN}. 

On the Reddit data sets and the COLLAB data set, $NetEmd_{G5}$ significantly outperforms other state-of-the-art graph classifiers. On the other hand, we find that $NetEmd_{G5}$ performs poorly on the IMDB data sets. This can be traced back to the large number of complete graphs present in the IMDB data sets: 139 out of the 1000 graphs in IMDB-Binary and 789 out of 1500 graphs in IMDB-Multi are complete graphs which correspond to ego-networks of actors having acted only in a single film. By definition, $NetEmd_{G5}$ cannot distinguish between complete graphs of different sizes since all graphlet degree distributions are concentrated on a single value in complete graphs. The spectral variant $NetEmd_S$ is not affected by this and we find that $NetEmd_S$ is either on par with or outperforms the other non-$NetEmd$ graph classifiers on all six data sets. 

We also tested $NetEmd$ on benchmark data sets representing chemical compounds and protein structures. Unlike the social network data sets, in these data sets nodes and edges are labeled to reflect domain specific knowledge such as atomic number, amino acid type and bond type. Although $NetEmd$, in contrast to the other graph kernels, does not rely on domain specific knowledge in the form of node or edge labels, we found that $NetEmd$ outperforms many of the considered graph kernels coming only second to the Weisfeiler-Lehman \citep{WL} type kernels in terms of overall performance (see Appendix \ref{MLC}).

\section{Discussion}
Starting from basic principles, we have introduced a general network comparison methodology, $NetEmd$, that is aimed at capturing common generating processes in networks. We tested $NetEmd$ in a large variety of experimental settings and found that $NetEmd$ successfully identifies similar networks at multiple scales even when networks differ significantly in terms of size and density, generally outperforming other graphlet based network comparison measures. Even when based only on graphlets up to size 3 (i.e. edges, 2-paths and triangles), $NetEmd$ has performance comparable to the state of the art, making $NetEmd$ feasible even for networks containing billions of edges and nodes.   

By exploring machine learning applications we showed that $NetEmd$ captures topological similarity in a way that relates to the function of networks and outperforms state-of-the art graph classifiers on several graph classification benchmarks.

Although we only considered variants of $NetEmd$ that are based on distributions of graphlets and spectra of Laplacian operators in this paper, $NetEmd$ can also be applied to other graph features in a straightforward manner. For instance, distributions of paths and centrality measures might capture larger scale properties of networks and their inclusion into $NetEmd$ might lead to a more refined measure. 

\section*{Data availability} The source code for $NetEmd$ is freely  available at:www.opig.ox.ac.uk/resources
\section*{Acknowledgements}This work was in part supported by  EPSRC grant EP/K032402/1 (A.W, G.R, C.D and R.G) and EPSRC grants EP/G037280/1 and EP/L016044/1 (C.D). L.O acknowledges the support of Colciencias through grant 568.  R.G. acknowledges support from the COST Action CA15109 and is currently supported by a Dame Kathleen Ollerenshaw Research Fellowship. C.D. and G.R. acknowledge the support of the Alan Turing Institute (grant EP/NS10129/1).

We thank Xiaochuan Xu and Martin O'Reilly for useful discussions.

\appendix
\section{Implementation}  

\subsection{Graphlet distributions.} In the main paper, both the graphlet degree distribution and graphlet counts in 1-step ego networks were used as inputs for $NetEmd$. 
\paragraph{Graphlet degree distributions}
The graphlet degree \citep{gdda} of a node  specifies the number of graphlets (small induced subgraphs) of a certain type the node appears in, while distinguishing between different positions the node can have in a graphlet. Different positions within a graphlet correspond to the orbits of the automorphism group of the graphlet. Among graphs on two to four nodes, there are 9 possible graphs and 15 possible orbits. Among graphs on two to five nodes there are 30 possible graphs and 73 possible orbits.

\paragraph{Graphlet distributions based on ego-networks.} Another way of obtaining graphlet distributions  is to consider graphlet counts in ego-networks \citep{2014waqar}. The $k$-step ego-network of a node $i$ is defined as the subgraph induced on all the nodes that can be reached from $i$ (including $i$) in less than $k$ steps. For a given $k$, the distribution of a graphlet $m$ in a network $G$ is then simply obtained by counting the occurrence of $m$ as an induced subgraph in the $k$-step ego-networks of each individual node.

\subsection{Step-wise implementation}
In this paper, for integer valued network features such as graphlet based distributions, we base our implementation on the probability distribution that corresponds to the histogram of feature $t$ with bin width 1 as $p_t(G)$. $NetEmd$ can also be defined on the basis of discrete empirical distributions i.e. distributions consisting of point masses (See Section \ref{pr}).

Here we summarise the calculation of the $NetEmd_T(G,G')$ distance between networks $G$ and $G'$ (with $N$ and $N'$ nodes respectively), based on the comparison of the set of local network features $T=\{t_1,\ldots,t_m\}$ of graphlet degrees corresponding to graphlets up to size $k$. 

\begin{enumerate}

\item First one computes the graphlet degree sequences corresponding to graphlets up to size $k$ for networks $G$ and $G'$. This can be done efficiently using the algorithm ORCA \citep{hovcevar2014combinatorial}. For the graphlet degree $t_1$ compute a histogram across all $N$ nodes of $G$ having bins of width 1 of which the centers are at their respective values. This histogram is then normalized to have total mass 1. We then interpret the histogram as the (piecewise continuous) probability density function of a  random variable. This probability density function is denoted by  $p_{t_1}(G)$.  The standard deviation of $p_{t_1}(G)$ is then computed, and is used to rescale the distribution so that it has variance 1. This distribution is denoted by $\widetilde{p_{t_1}(G)}$.

\item Repeat the above step for network $G'$, and denote the resulting distribution by $\widetilde{p_{t_1}(G')}$.  Now compute
\begin{equation*}
NetEmd_{t_1}^*(G,G')=\mathrm{inf}_{c\in\mathbb{R}}\big(EMD\big(\widetilde{p_{t_1}(G)}(\cdot+c),\widetilde{p_{t_1}(G')}(\cdot)\big)\big).
\end{equation*}
In practice, this minimisation over $c$ is computed using a suitable optimization algorithm. In our implementation we use the Brent-Dekker algorithm \citep{brent1971algorithm} with an error tolerance of 0.00001 and with the number of iterations upper bounded by 150.  

\item Repeat the above two steps for the network features $t_2,\ldots,t_m$ and compute
\begin{equation*}
NetEmd_T(G,G')=\frac{1}{m}\sum_{j=1}^m NetEmd_{t_j}^*(G,G').
\end{equation*}

\end{enumerate}

\subsection{Example: $EMD^*$ for Gaussian distributions}
Suppose that $p$ and $q$ are $N(\mu_1,\sigma_1^2)$ and $N(\mu_2,\sigma_2^2)$ distributions, respectively.  Then

\begin{align*}
EMD^*(p,q) &=\inf_{c\in\mathbb{R}}\Big(EMD\big(\tilde{p}(\cdot+c),\tilde{q}(\cdot)\big)\Big)\\
&=EMD \Big(\tilde{p}(\cdot-\frac{\mu_1}{\sigma_1}+\frac{\mu_2}{\sigma_2}),\tilde{q}(\cdot)\Big)\\
&=EMD\big(\tilde{q}(\cdot),\tilde{q}(\cdot)\big)=0.
\end{align*}

Here we used that if $X\sim N(\mu_1,\sigma_1^2)$ and $Y\sim N(\mu_2,\sigma_2^2)$, then $\frac{X}{\sigma_1}+c\sim N(\frac{\mu_1}{\sigma_1}-c,1)$ and $\frac{Y}{\sigma_2}\sim N(\frac{\mu_2}{\sigma_2},1)$, and these two distributions are equal if $c=\frac{\mu_1}{\sigma_1}-\frac{\mu_2}{\sigma_2}$.

\subsection{Spectral NetEmd}
When using spectra of graph operators, which take real values instead of the integer values one has in the case of graphlet distributions, we use the empirical distribution consisting of point masses for computing $NetEmd$. For more details see Section \ref{pr} of this appendix. 
\subsection{Computational complexity}
The computational complexity of graphlet based comparison methods is dominated by the complexity of enumerating graphlets. For a network of size $N$ and maximum degree $d$,  enumerating all connected graphlets up to size $m$  has complexity $O(Nd^{m-1})$, while counting all graphlets up to size $m$ in all $k$-step ego-networks has complexity $O(Nd^{k+m-1})$. Because most real world networks are sparse, graphlet enumeration algorithms tends to scale more favourably in practice than the worst case upper bounds given above.  

In the case of spectral measures, the most commonly used algorithms for computing the eigenvalue spectrum have complexity $O(N^3)$. Recent results show that the spectra of graph operators can be approximated efficiently in $O(N^2)$ time \citep{thune2013eigenvalues}. 

Given the distribution of a feature $t$, computing $EMD_t^*(G,G')$ has complexity $O(k(s+s')\mathrm{log}(s+s'))$, where $s$ and $s'$ are the number of different values $t$ takes in $G$ and $G'$ respectively and $k$ is the maximum number function calls of the optimization algorithm used to align the distributions. For node based features such as motif distributions, the worst case complexity is $O(k(N(G)+N(G'))\mathrm{log}(N(G)+N(G')))$, where $N(G)$ is the number of nodes of $G$, since the number of different values $t$ can take is bounded by the number of nodes.

\section{Proof that $NetEmd$ is a distance measure}\label{metric}
We begin by stating a definition.  A \emph{pseudometric} on a set $X$ is a non-negative real-valued function $d:X\times X\rightarrow [0,\infty)$ such that, for all $x,y,z\in X$,
\begin{enumerate}

\item  $d(x,x)= 0$;

\item  $d(x,y)=d(y,x)$ (symmetry);

\item  $d(x,z)\leq d(x,y)+d(y,z)$ (triangle inequality).

\end{enumerate}

If Condition 1 is replaced by the condition that $d(x,y)=0\iff x=y$ then $d$ defines a \emph{metric}.  Note that this requirement can only be satisfied by a network comparison measure that is based on a complete set of graph invariants and hence network comparison measures in general will not satisfy this requirement.  

\paragraph{Proposition}Let $M$ denote the space of all real-valued probability measures supported on $\mathbb{R}$ with finite, non-zero variance.  Then the $EMD^*$ distance between probability measures, $\mu_X$ and $\mu_Y$ in $M$ defined by
\begin{equation*}EMD^*(\mu_X,\mu_Y)=\inf_{c\in\mathbb{R}}EMD(\tilde{\mu}_X(\cdot),\tilde{\mu}_Y(\cdot+c)),
\end{equation*}
defines a pseudometric on the space of probability measures $M$.

\paragraph{Proof}  We first note that if $\mu_X\in M$ then $\tilde{\mu}_X(\cdot+c)\in M$ for any $c\in\mathbb{R}$.  Let us now verify that $EMD^*$ satisfies all properties of a pseudometric.  Clearly, for any $\mu_X\in M$, we have $0\leq EMD^*(\mu_X,\mu_X)\leq EMD(\tilde{\mu}_X(\cdot),\tilde{\mu}_X(\cdot))=0$, and so $EMD^*(\mu_X,\mu_X)=0$.  Symmetry holds, since for, any $\mu_X$ and $\mu_Y$ in $M$,
\begin{align*}EMD^*(\mu_X,\mu_Y)&=\inf_{c\in\mathbb{R}}EMD(\tilde{\mu}_X(\cdot),\tilde{\mu}_Y(\cdot+c))\\&=\inf_{c\in\mathbb{R}}EMD(\tilde{\mu}_Y(\cdot+c),\tilde{\mu}_X(\cdot))\\
&=\inf_{c\in\mathbb{R}}EMD(\tilde{\mu}_Y(\cdot),\tilde{\mu}_X(\cdot+c))\\&=EMD^*(\mu_Y,\mu_X).
\end{align*}
Finally, we verify that $EMD^*$ satisfies the triangle inequality. Suppose $\mu_X$, $\mu_Y$ and $\mu_Z$ are probability measures from the space $M$, then so are $\tilde{\mu}_X(\cdot+a)$, $\tilde{\mu}_Y(\cdot+b)$ for any $a,b\in\mathbb{R}$. Since EMD satisfies the triangle inequality, we have, for any $a,b\in\mathbb{R}$,
\begin{equation*}\label{emdeqn1}EMD(\tilde{\mu}_X(\cdot+a),\tilde{\mu}_Y(\cdot+b))\leq EMD(\tilde{\mu}_X(\cdot+a),\tilde{\mu}_Z(\cdot))+EMD(\tilde{\mu}_Y(\cdot+b),\tilde{\mu}_Z(\cdot)).
\end{equation*}
Since the above inequality holds for all $a,b\in\mathbb{R}$, we have that
\begin{align*}EMD^*&(\mu_X,\mu_Y)=\inf_{c\in\mathbb{R}}EMD(\tilde{\mu}_X(\cdot+c),\tilde{\mu}_Y(\cdot))\\
&=\inf_{a,b\in\mathbb{R}}EMD(\tilde{\mu}_X(\cdot+a),\tilde{\mu}_Y(\cdot+b)) \\
&\leq \inf_{a,b\in\mathbb{R}}\big[EMD(\tilde{\mu}_X(\cdot+a),\tilde{\mu}_Z(\cdot))+EMD(\tilde{\mu}_Y(\cdot+b),\mu_Z(\cdot))\big] \\
&=\inf_{a\in\mathbb{R}}\big[EMD(\tilde{\mu}_X(\cdot+a),\tilde{\mu}_Z(\cdot))+\inf_{b\in\mathbb{R}}EMD(\tilde{\mu}_Y(\cdot+b),\tilde{\mu}_Z(\cdot))\big] \\
&=\inf_{a\in\mathbb{R}}EMD(\tilde{\mu}_X(\cdot+a),\tilde{\mu}_Z(\cdot))+\inf_{b\in\mathbb{R}}EMD(\tilde{\mu}_Y(\cdot+b),\tilde{\mu}_Z(\cdot))\\
&=EMD^*(\mu_X,\mu_Z)+EMD^*(\mu_Y,\mu_Z),
\end{align*}
as required.  We have thus verified that $EMD^*$ satisfies all properties of a pseudometric. \hfill $\Box$

\section{Generalization of $EMD^*$ to point masses}\label{pr}
Although in the case of graphlet based features we based our implementation of $NetEmd$ on probability distribution functions that correspond to normalized histograms havning bin width 1 $NetEmd$ can also be based on empirical distributions consisting of collections of point masses located at the observed values. 


The definition of $EMD^*$ can be generalized to include distributions of zero variance, i.e. unit point masses. Mathematically, the distribution of a point mass at $x_0$ is given by the Dirac measure $\delta_x(x_0)$. Such distributions are frequently encountered in practice since some graphlets do not occur in certain networks. 

First, we note that unit point masses are always mapped onto unit point masses under rescaling operations. Moreover, for a unit point mass $\delta_x(x_0)$ we have that $\mathrm{inf}_{c\in\mathbb{R}}(EMD(\tilde{p}(\cdot+c),\delta_x(x_0)))$ $=\mathrm{inf}_{c\in\mathbb{R}}\left(EMD(\tilde{p}(\cdot+c),\delta_x(kx_0))\right)$  for all $p\in M$ and $k>0$. Consequently, $EMD^*$ can be generalized to include unit point masses in a consistent fashion by always rescaling them by 1: 
\begin{equation*}\label{emdmet2}
EMD^*(p,q)=\mathrm{inf}_{c\in\mathbb{R}}\big(EMD(\hat{p}(\cdot+c),\hat{q})\big),
\end{equation*}
where $\hat{p}=\tilde{p}$ (as in Eq. \ref{emdmet}) if $p$ has a non-zero variance, and $\hat{p}=p$ if $p$ has variance zero.

\section{Sub-sampling}
$NetEmd$ is well suited for the sub-sampling procedure from \citep{SS}. Following this procedure we base the graphlet distributions used as an input of $NetEmd$ on a sample of nodes rather than the whole network. 

\begin{figure*}[!htb]
\centering 
\subfloat[Synthetic networks]{\includegraphics[width=0.45\linewidth]{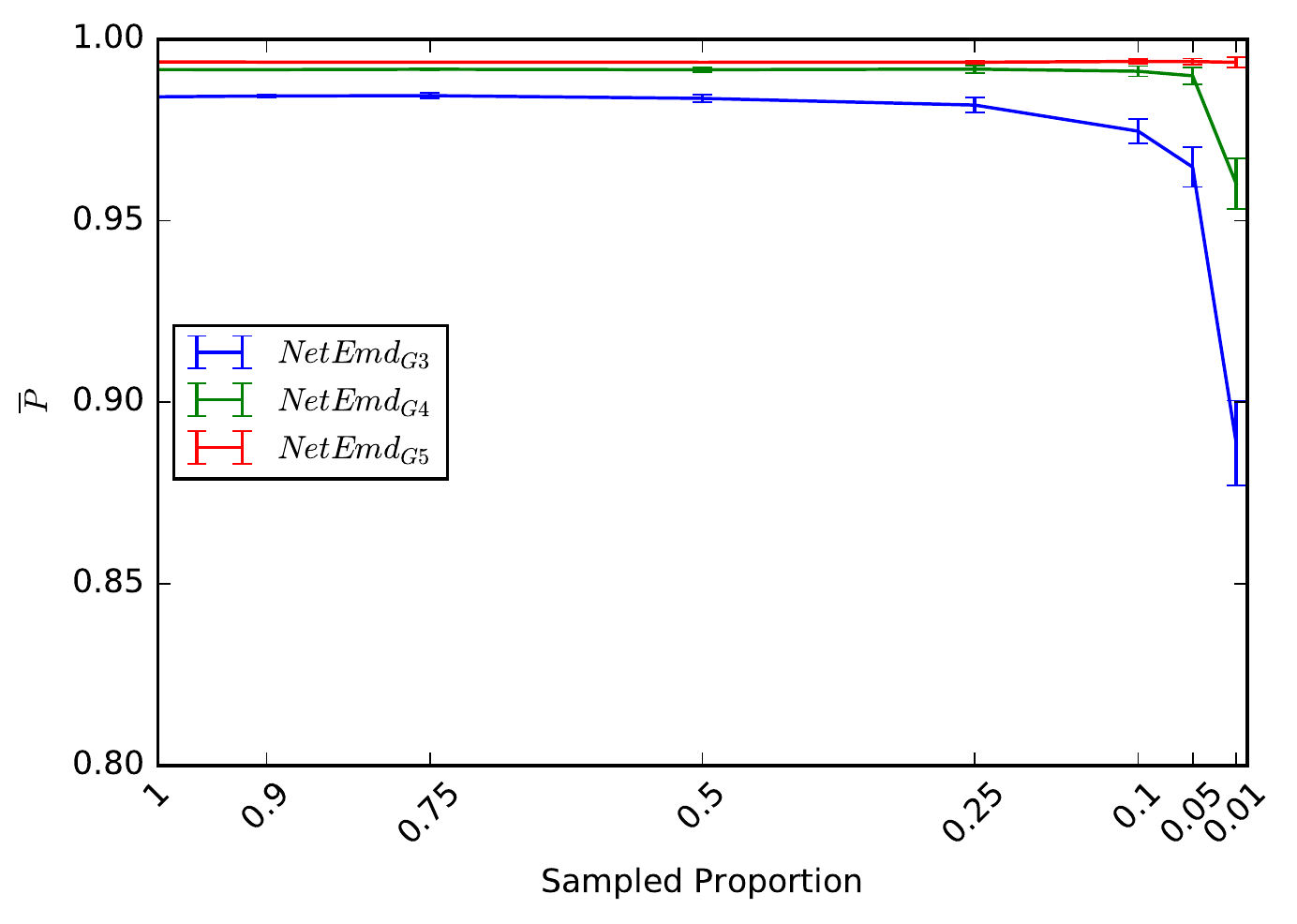}}
\qquad
\subfloat[Onnela et al.]{\includegraphics[width=0.45\linewidth]{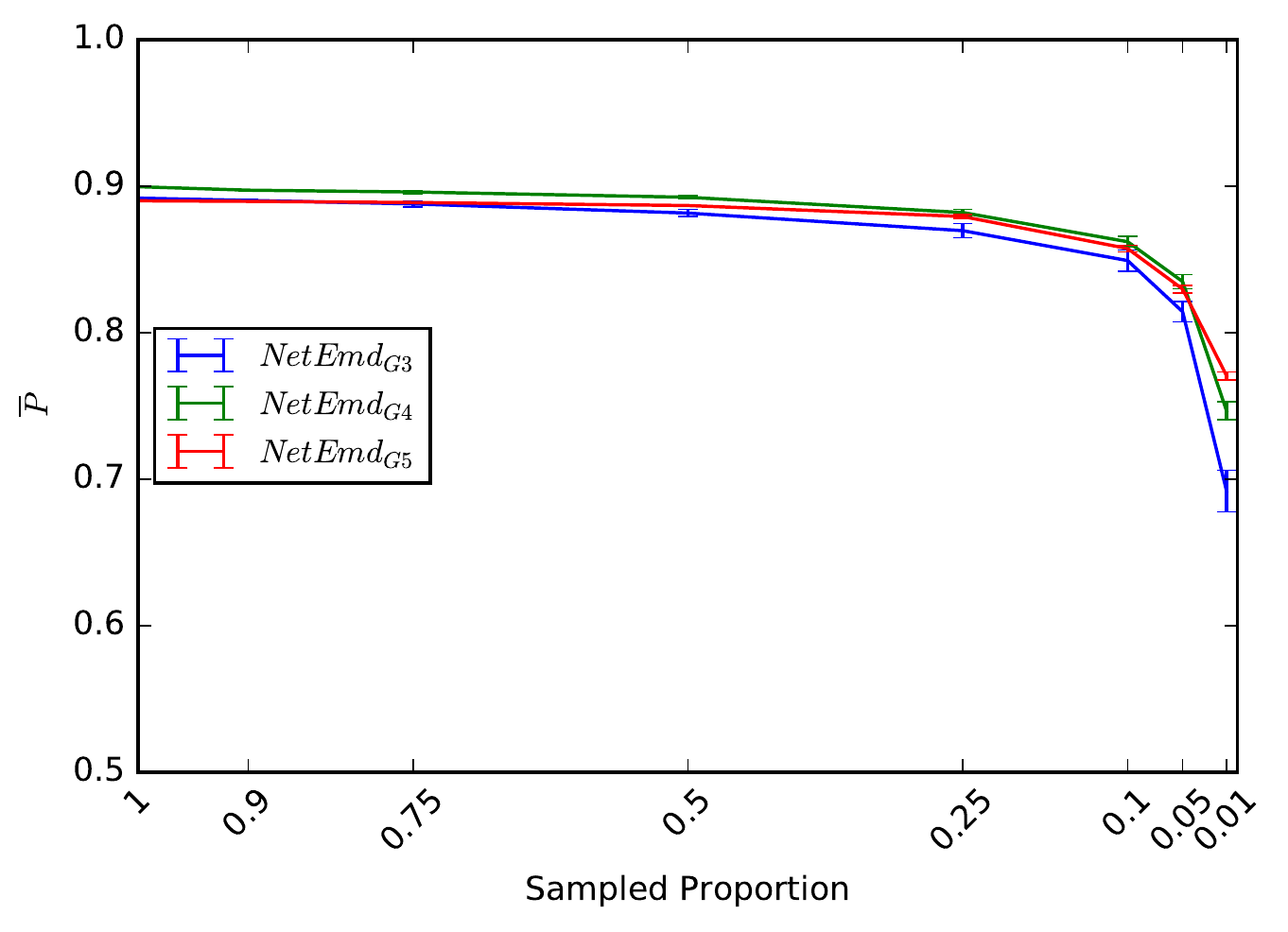}}
\caption{The $\overline{P}$ values for different variants of $NetEmd$ under sub-sampling for a) a set of 80 synthetic networks coming from eight different random graph models with 2500 nodes and average degree 20, b) for the Onnela et al. data set showing the average and standard deviation over 50 experiments for each sampled proportion. Note that the performance of $NetEmd$ under sub-sampling is remarkably stable and is close to optimal even when only $10\%$ of nodes are sampled. For synthetic networks we find that the stability of $NetEmd$ increases as the size of the graphlets used in the input is increased.}
\label{SS}
\end{figure*}

Figure \ref{SS} shows the $\overline{P}$ scores for variants of $NetEmd$ on a set of synthetic networks and the Onnela et al. data set.  We find that the performance of $NetEmd$ is stable under sub-sampling and that in general using a sample of only $10\%$ of the nodes produces results comparable to the case where all nodes are used.   
\begin{figure*}[!htb]
 \centering
 \subfloat[Heatmap of $NetEmd_{G5}$ for $RG_3$.]{\includegraphics[width=0.42\linewidth]{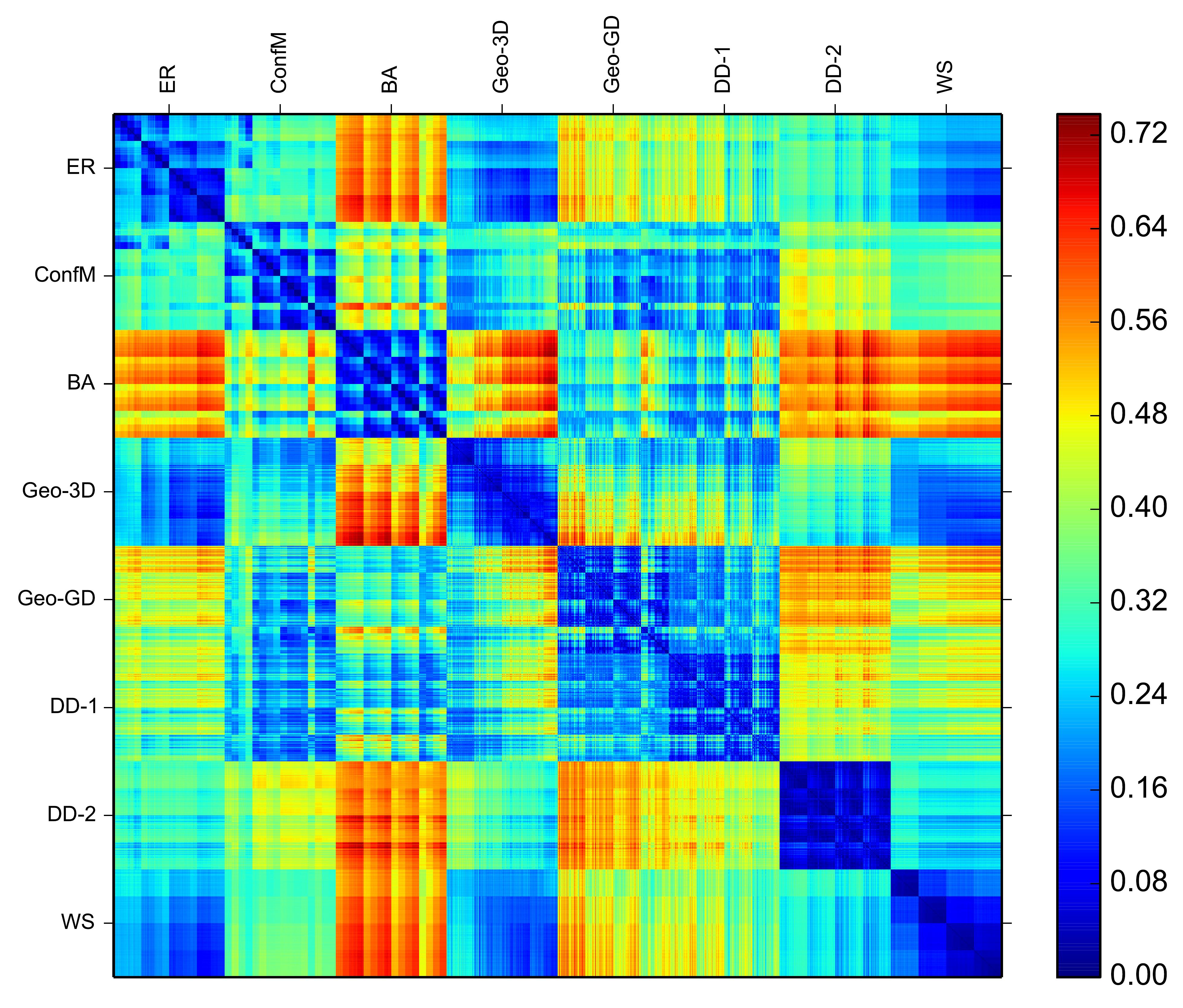}
}\quad
\subfloat[Heatmap of $GCD11$ for $RG_3$.]{\includegraphics[width=0.42\linewidth]{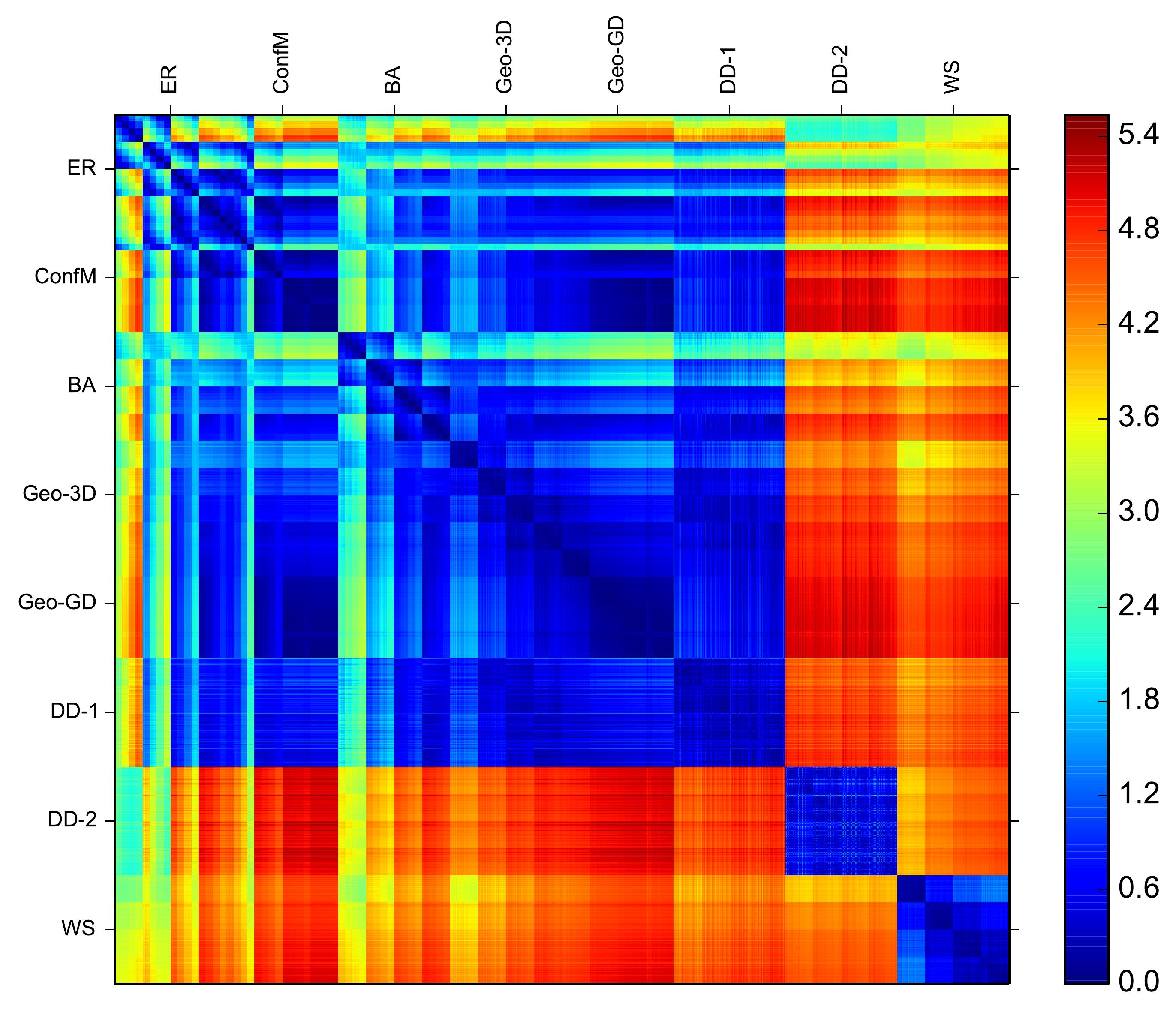}}
\caption{ a) and b) show the heatmaps of pairwise distances on $RG_3$ ($N\in\{1250,2500,5000,10000\}$ and $k\in\{10,20,40,80\}$) according to  $NetEmd_{G5}$ and next best performing measure $GCD11$, respectively. In the heat map, networks are ordered from top to bottom in the following order: model, average degree and node count. Although we observe some degree of off diagonal mixing the heatmap of $NetEmd$ still shows 8 diagonal blocks corresponding to different generative models in contrast to the heat map of $GCD11$.}
\label{fig:RG3}
\end{figure*}

\section{Results for data sets of chemical compounds and proteins}\label{MLC}

We also tested $NetEmd$ on benchmark data sets representing chemical compounds (MUTAG, NCI1 and NCI109) and protein structures (ENZYMES and D\&D). MUTAG \citep{mutag} is a data set of 188 chemical compounds that are labelled according to their mutagenic effect on Salmonella typhimurium. NCI1 and NCI109 represent sets of chemical compounds which are labelled for their activity against non-small cell lung cancer and ovarian cancer cell lines, respectively \citep{NCI}. Nodes and edges in MUTAG, NCI1 and NCI109 are labeled by atomic number and bound type, respectively. ENZYMES and D\&D \citep{borgpro} consist of networks representing protein structures at the level of tertiary structure and amino acids respectively. While networks in ENZYMES are classified into six different enzyme classes, networks in D\&D are classified according to whether or not they correspond to an enzyme. Nodes in ENZYMES are labelled according to structural element type and according to amino acid types in D\&D. 

Classification accuracies obtained using $NetEmd$ on the data sets of chemical compounds and protein structures are given in Table \ref{CSVM}, along with results for other graph kernels reported in \citep{WL}. For a detailed description of these kernels we refer to \citep{WL} and the references therein. Note that, in contrast to all other kernels in Table \ref{CSVM}, $NetEmd$ does not use any domain specific knowledge in the form of node or edge labels. Node and edge labels are highly informative for all five classification tasks - as shown in \citep{halting}.

\begin{table}[!t]

\resizebox{\textwidth}{!}{\begin{tabular}[c,width=\linewidth]{lccccc}
Kernel&MUTAG& NCI1& NCI109 & ENZYMES& D \& D\\
\midrule

$NetEmd_{G5}$ &83.71 $\pm$1.16&78.59$\pm$0.28&76.71$\pm$0.34&46.55$\pm$1.25&78.01 $\pm$0.38 \\
$NetEmd_{S}$ &83.30 $\pm$1.20&77.36$\pm$0.38&76.14$\pm$0.27&42.75$\pm$0.78&76.74 $\pm$0.43 \\

WL subtree&82.05$\pm$0.36&82.19 $\pm$0.18&82.46 $\pm$0.24&52.22$\pm$1.26&79.78 $\pm$0.36\\

WL edge &81.06$\pm$1.95&84.37$\pm$0.30&84.49$\pm$0.20&53.17$\pm$2.04&77.95$\pm$0.70\\

WL shortest path&83.78$\pm$1.46&84.55$\pm$0.36&83.53$\pm$0.30&59.05$\pm$1.05&79.43$\pm$0.55\\

Ramon \& G\"artner&85.72$\pm$0.49&61.86$\pm$0.27&61.67$\pm$0.21&13.35$\pm$0.87&57.27$\pm$0.07\\

p-random walk&79.19$\pm$1.09&58.66$\pm$0.28&58.36$\pm$0.94&27.67$\pm$0.95&66.64$\pm$0.83\\

Random Walk&80.72$\pm$0.38&64.34$\pm$0.27&63.51$\pm$0.18&21.68$\pm$0.94&71.70$\pm$0.47\\

Graphlet count&75.61$\pm$0.49&66.00$\pm$0.07&66.59$\pm$0.08&32.70$\pm$1.20&78.59$\pm$0.12\\

Shortest path&87.28$\pm$0.55&73.47$\pm$0.11&73.07$\pm$0.11&41.68$\pm$1.79&78.45$\pm$0.26\\

\bottomrule
\end{tabular}}
\caption{10 fold cross validation accuracies of Gaussian kernels based on $NetEmd_{G5}$ and $NetEmd_S$ and other kernels reported in \citep{WL}.}

\label{CSVM}
\end{table}
On MUTAG, $NetEmd$ achieves an accuracy that is comparable to the Weisfeiler-Lehman (WL) shortest path kernel, but is outperformed by the shortest path kernel and the kernel by Ramon \& G\"artner. While on NCI1, NCI109 and ENZYMES, $NetEmd$ is outperformed only by WL kernels, on D\&D $NetEmd$ achieves a classification accuracy that is comparable to the best performing kernels. Notably, on D\&D $NetEmd$ also outperforms the vector model by Dobson and Doig \citep{DD} (classification accuracy: 76.86$\pm$1.23) which is based on 52 physical and chemical features without using domain specific knowledge i.e. solely based on graph topology.
\subsection{Implementation of C-SVMs}
 Following the procedure in \citep{WL} we use  10-fold  cross  validation  with  a C-SVM \citep{csvm} to  test  classification  performance. We use the python package scikit-learn \citep{pedregosa2011scikit} which is based is build on libsvm implementation \citep{chang2011libsvm}. The $C-value$  of the C-SVM and the $\alpha$ for the Gaussian kernel is tuned independently for each fold using training data from that fold only. Each experiment is repeated 10 times, and average prediction accuracies and their standard deviations are reported.

We also note that note for all values of $\alpha$ is the Gaussian NetEmd kernel is positive semidefinite (psd)  \citep{jayasumana2015kernel}. The implication is that the C-SVM converges to a stationary point that is not always guaranteed to be global optimum. Although there exist alternative algorithms \citep{luss2008support} for training C-SVMs with indefinite kernels which might result in better classification accuracy, here we chose to use the standard libsvm-algorithm in order to ensure a fair comparison between kernels. For a discussion of support vector machines with indefinite kernels see \citep{haasdonk2005feature}.

\section{ Detailed description of data sets and models}
\subsection{Synthetic networks and random graph models}\label{models}
$RG_1$ consists of 16 sub data sets corresponding to combinations of $N\in\{1250,2500,5000,10000\}$ and $k\in\{10,20,40,80\}$ containing 10 realizations for each model i.e. contain 80 networks each. 

In $RG_2$ the size $N$ and average degree $k$ are increased independently in linear steps to twice their initial value ($N\in\{2000,3000,4000\}$ and $k\in\{20,24,28,32,36,40\}$) and contains 10 realizations per model parameter combination, resulting in a data set of $3\times6\times8\times10=1440$ networks. 

In $RG_3$ the size $N$ and average degree $k$ are increased independently in multiples of 2 to 8 times their initial value ($N\in\{1250,2500,5000,10000\}$ and $k\in\{10,20,40,80\}$) and again contains 10 realizations per model parameter combination, resulting in a data set of $4\times4\times8\times10=1280$ networks.  The models are as follows.

\subsubsection{The Erd\H{o}s-R{\'e}nyi model}
We consider the  Erd\H{o}s-R{\'e}nyi (ER) model \citep{1960er} $G(N,m)$ where $N$ is the number of nodes and $m$ is the number of edges. The edges are chosen uniformly at random without replacement from the $\binom{N}{2}$ possible edges. 
\subsubsection{The configuration model}
Given a graphical degree sequence, the configuration model creates a random graph that is drawn uniformly at random from the space of all graphs with the given degree sequence. The degree sequence of the configuration models used in the paper is taken to be degree sequence of a duplication divergence model that has the desired average degree. 

\subsubsection{The Barab{\'a}si Albert preferential attachment model}
In the Barab{\'a}si-Albert model \citep{barabasi1999emergence} a network is generated starting from a small initial network to which nodes of degree $m$ are added iteratively and the probability of connecting the new node to an existing node is proportional to the degree of the existing node. 

\subsubsection{Geometric random graphs}
Geometric random graphs \citep{1961gilbert} are constructed under the assumption that the nodes in the network are embedded into a $D$ dimensional space, and the presence of an edge depends only on the distance between the nodes and a given threshold $r$. The model is constructed by placing $N$ nodes uniformly at random in an $D$-dimensional square $[0,1]^D$. Then edges are placed between any pair of nodes for which the distance between them is less or equal to the threshold $r$. We use $D=3$ and set $r$ to be the threshold that results in a network with the desired average degree, while the distance is the Euclidean distance.  
\subsubsection{The geometric gene duplication model}
The geometric gene duplication model is a geometric model \citep{2008higham} in which
the nodes are distributed in 3 dimensional Euclidean space $\mathbb{R}^3$ according to the following rule. Starting from an small initial set of nodes in three dimensions, at each step a randomly chosen node is selected and a new node is placed at random within a Euclidean distance $d$ of this node. The process is repeated until the desired number of nodes is reached. Nodes within a certain distance $r$ are then connected. We fix $r$ to obtain the desired average degree. 

\subsubsection{The duplication divergence model of V\'azquez et al.}
The duplication divergence model of V\'azquez et al. \citep{DD1} is defined by the following growing rules: (1) Duplication: A node $v_i$ is randomly selected and duplicated ($v_i'$) along with all of its interactions. An edge between $v_i$ and $v_i'$ is placed with probability $p$. (2) Divergence: For each pair of duplicated edges $\{(v_i,v_k);(v_i',v_k)\}$; one of the duplicated edges is selected uniformly at random and then deleted with probability $q$. This process is followed until the desired number of nodes is reached. In our case we fix $p$ to be 0.05 and adjust $q$ through a grid search to obtain a network that on average has the desired average degree. 
\subsubsection{The duplication divergence of Ispolatov et al.} The duplication divergence model of Ispolatov et al. \citep{DD2} starts with an initial network consisting of a single edge and then at each step a random node is chosen for duplication and the duplicate is connected to each of the neighbours of its parent with probability $p$.
We adjust $p$ to obtain networks that have on average the desired average degree. 

\subsubsection{The Watts-Strogatz model}
The Watts-Strogatz model, \citep{1998watts} creates graphs that interpolate between regular graphs and ER graphs. The model starts with a ring of $n$ nodes in which each node is connected to its $k$-nearest neighbours in both directions of the ring. Each edges is rewired with probability $p$ to a node which is selected uniformly at random. While $k$ is adjusted to obtain networks having the desired average degree we take $p$ to be 0.05.

\subsection{Real world data sets}

Summary statistics of the data sets are given in Table \ref{RW}.
\begin{table}[!t]

\centering
\resizebox{\textwidth}{!}{
\begin{tabular}[c]{lcccccccccc}
Data set&\#Networks&$N_{min}$&Median($N$) &$N_{max}$&$E_{min}$&Median($E$) &$E_{max}$&$d_{min}$&Median($d$) &$d_{max}$\\
\midrule
RWN&167&24&351&62586&76 &2595& 824617&7.55e-05& 0.0163 &0.625
\\

Onnela et al.&151&30 &918& 11586&62 &2436&232794&4.26e-5& 0.0147&0.499
\\
\midrule

AS-caida&122& 8020 &22883& 26475 &18203 &46290& 53601&1.48e-4& 1.78e-4& 5.66e-4
\\
AS-733&732& 493 &4180.5& 6474 &1234& 8380.5& 13895&6.63e-4& 9.71e-4& 1.01e-2\\
World Trade Networks&53 &156 &195& 242& 5132 &7675& 18083&0.333& 0.515&0.625\\
\midrule
Reddit-Binary&2000&6& 304.5&3782
&4 &379& 4071
&5.69e-4 &8.25e-3& 0.286
\\
Reddit-Multi-5k&4999 &22 &374& 3648&
21& 422& 4783
&6.55e-4&6.03e-3 &0.091
\\
Reddit-Multi-12k&11929&2 &280& 3782
&1 &323& 5171&5.69e-4&8.27e-3& 1.0
\\
COLLAB&5000 &32 &52& 492&
60 &654.5 &40120&
0.029& 0.424& 1.0
\\
IMDB-Binary&1000 
&12 &17 &136
&26 &65& 1249&0.095& 0.462 &1.0
\\
IMDB-Multi&1500&7& 10& 89&12& 36&1467&
0.127& 1.0 &1.0\\
\midrule
MUTAG&188 &10 &17.5& 28 &10 &19& 33&0.082 &0.132& 0.222\\
NCI1&4110 &3 &27&111& 2 &29& 119&0.0192& 0.0855& 0.667
\\
NCI109&4127 &4 &26& 111& 3 &29& 119&0.0192& 0.0862&0.5\\
ENZYMES&600& 2 &32 &125 &1 &60 &149&0.0182& 0.130& 1.0\\
D\&D&1178 &30 &241& 5748 &63& 610.5 &14267&8.64e-4& 0.0207& 0.2
\\
\bottomrule
\end{tabular}}
\caption{Summary statistics of data sets $N$, $E$ and $d$ stand for the number of nodes, number of edges and edge density, respectively.}
\label{RW}
\end{table}

\subsubsection{Real world networks from different classes (RWN)}
We compiled a data set consisting of 10 different classes of real world networks: social networks, metabolic networks, protein interaction networks, protein structure networks, food webs, autonomous systems networks of the internet, world trade networks,  airline networks, peer to peer file sharing networks and scientific collaboration networks. Although in some instances larger versions of these data sets are available, we restrict the maximum number of networks in a certain class to 20 by taking random samples of larger data sets in order to avoid scores being dominated by larger network classes.  

The class of social networks consists of 10 social networks from the Pajek data set which can be found at http://vlado.fmf.uni-lj.si/pub/networks/data/default.htm (June 12th 2015) (Networks: 'bkfrat', 'bkham', 'bkoff', 'bktec', 'dolphins', 'kaptailS1', 'kaptailS2', 'kaptailT1', 'kaptailT2', 'karate', 'lesmis', 'prison') and a sample of 10 Facebook networks from  \citep{traud2012social} (Networks:'Auburn71', 'Bucknell39', 'Caltech36', 'Duke14', 'Harvard1', 'JMU79', 'MU78', 'Maine59', 'Maryland58', 'Rice31', 'Rutgers89', 'Santa74', 'UC61', 'UC64', 'UCLA26', 'UPenn7', 'UVA16', 'Vassar85', 'WashU32', 'Yale4'). The class of metabolic networks consists of 20 networks taken \citep{jeong2000large} (Networks: 'AB', 'AG', 'AP', 'AT', 'BS', 'CE', 'CT', 'EF', 'HI', 'MG', 'MJ', 'ML', 'NG', 'OS', 'PA', 'PN', 'RP', 'TH', 'TM', 'YP'). The class of protein interaction networks consists of 6 networks from BIOGRID \citep{stark2006biogrid} (Arabidopsis thaliana, Caenorhabditis elegans,
Drosophila melanogaster, Homo sapiens,
Mus musculus and Saccharomyces cerevisiae downloaded: October 2015) and 5 networks  from HINT \citep{das2012hint} (Arabidopsis thaliana, Caenorhabditis elegans,
Drosophila melanogaster, Homo sapiens and 
Mus musculus (Version: June 1 2014))  and the protein interaction network of Echeria coli by Rajagopala et al. \citep{rajagopala2014binary}. The class of protein structure networks consists of a sample of 20 networks from the data set D\&D (Networks: 20, 119, 231, 279, 335, 354, 355, 369, 386, 462, 523, 529, 597, 748, 833, 866, 990, 1043, 1113, 1157). The class of food webs consists of 20 food webs from the Pajek data set: http://vlado.fmf.uni-lj.si/pub/networks/data/default.htm (June 10th 2015) (Networks: 'ChesLower', 'ChesMiddle', 'ChesUpper', 'Chesapeake', 'CrystalC', 'CrystalD', 'Everglades', 'Florida', 'Michigan', 'Mondego', 'Narragan', 'StMarks', 'baydry', 'baywet', 'cypdry', 'cypwet', 'gramdry', 'gramwet', 'mangdry', 'mangwet'). The class of internet networks consists of 10 randomly chosen networks from AS-733 \citep{as} 
(Networks:'1997/11/12',  '1997/12/28',  '1998/01/01', '1998/06/06', '1998/08/13', '1998/12/04', '1999/03/30', '1999/04/17',  '1999 /06/18',  '1999/08/30') and 10 randomly chosen networks from AS-caida \citep{as} (Networks: '2004/10/04',  '2006/01/23', '2006/03/27', '2006/07/10', '2006/09/25',  '2006/11/27', '2007/01/15', '2007/04/30',  '2007/05/28', '2007/09/24'). Both datasets are from SNAP \citep{snapnets}(June 1 2016).  The class of world trade networks is a sample of 20 networks of the larger data set considered in \citep{feenstra2005world,comtrade} (Networks: 1968, 1971, 1974, 1975, 1976, 1978, 1980, 1984, 1989, 1992, 1993, 1996, 1998, 2001, 2003, 2005, 2007, 2010, 2011, 2012). The airline networks were derived from the data available at: http://openflights.org/ (June 12 2015). For this we considered the 50 largest airlines from the database in terms of the number of destinations that the airline serves. For each airline a network is obtained by the considering all airports that are serviced by the airlines which are connected whenever there is direct flight between a pair of nodes. We then took a sample of 20 networks from this larger data set (Airline codes of the networks: 'AD', 'AF', 'AM', 'BA', 'DY', 'FL', 'FR', 'JJ', 'JL', 'MH', 'MU', 'NH', 'QF', 'SU', 'SV', 'U2', 'UA', 'US', 'VY', 'ZH'). The class of peer to peer networks consist of 9 networks of the Gnutella file sharing platform measured at different dates which are available at \citep{snapnets}. The scientific collaboration networks consists of 5 networks representing different scientific disciplines which were obtained from \citep{snapnets} (June 1 2015).

\subsubsection{Onnela et al. data set}
The Onnela et al. data set consists of all undirected and unweighted networks from the larger collection analysed in \citep{onnela}.
A complete list of networks and  class membership can be found in the supplementary information of \citep{2014waqar}.
\subsubsection{Time ordered data sets}
The data sets AS-caida and AS-733 each represent the internet measured at the level of autonomous systems at various points in time. Both data sets were downloaded from \citep{snapnets}(June 1 2015).

The World Trade Networks data set is based on the data set  \citep{feenstra2005world} for the years 1962-2000 and on UN COMTRADE \citep{comtrade} for the years 2001-2015. Two countries are connected in the network whenever they import or export a commodity from a each other within the given calendar year. The complete data set was downloaded from : http://atlas.media.mit.edu/en/resources/data/ on July 12 2015. 

\subsubsection{Machine learning benchmarks}

A short description of the social networks datasets was given in the main text. A more detailed description can be found in \citep{yanardag2015deep}. The social network data sets were downloaded from https://ls11-www.cs.tu-dortmund.de/staff/morris/graphkerneldatasets on September 2 2016. 

A short short description of the chemical compound and protein structure data sets was given in Section \ref{MLC}. A more detailed description of the data set can be found in \citep{WL}. These data sets were downloaded from: https://www.bsse.ethz.ch/mlcb/research/machine-learning/graph-kernels.html on June 12 2016.

\section{Performance evaluation via area under precision recall curve ?}

The area under precision recall curve (AUPRC) was used as a performance metric for network comparison measures by Yaveroglu et al. \citep{2014yaveroglu}. The AUPRC is based on a classifier that for a given distance threshold $\epsilon$ classifies pairs of networks to be similar whenever $d(G,G')<\epsilon$. 

A pair satisfying $d(G,G')<\epsilon$ is taken to be a true positive whenever  $G$ and $G'$ are from the same class. The AUPRC is then defined to be the area under the precision recall curve obtained by varying $\epsilon$ in small increments. However, AUPRC is problematic, especially in settings where one has more than two classes and when classes are separated at different scales. 

\begin{figure*}[!htb]
\subfloat[$d_1$]{\includegraphics[width=0.32\linewidth]{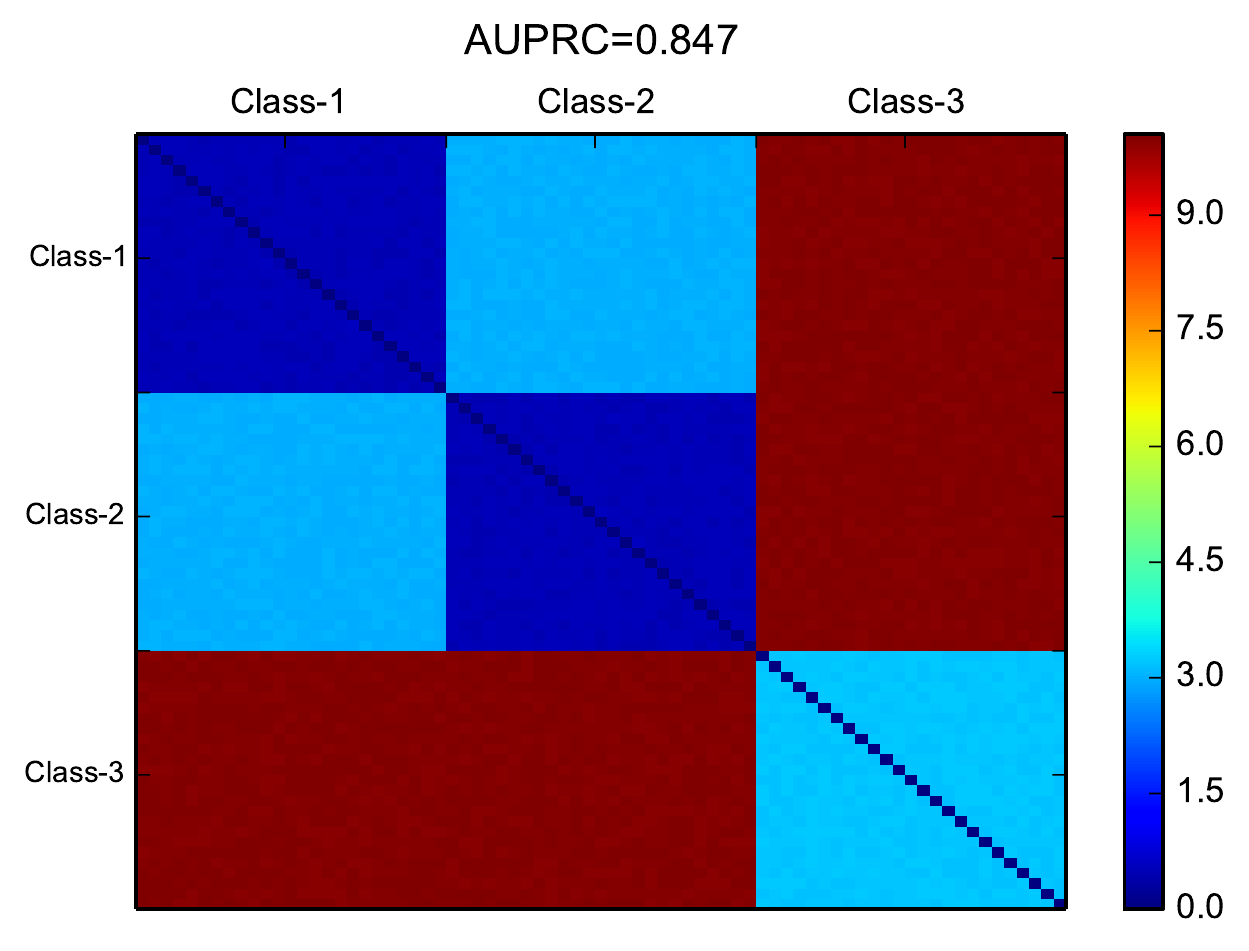}}
\subfloat[$d_2$]{\includegraphics[width=0.32\linewidth]{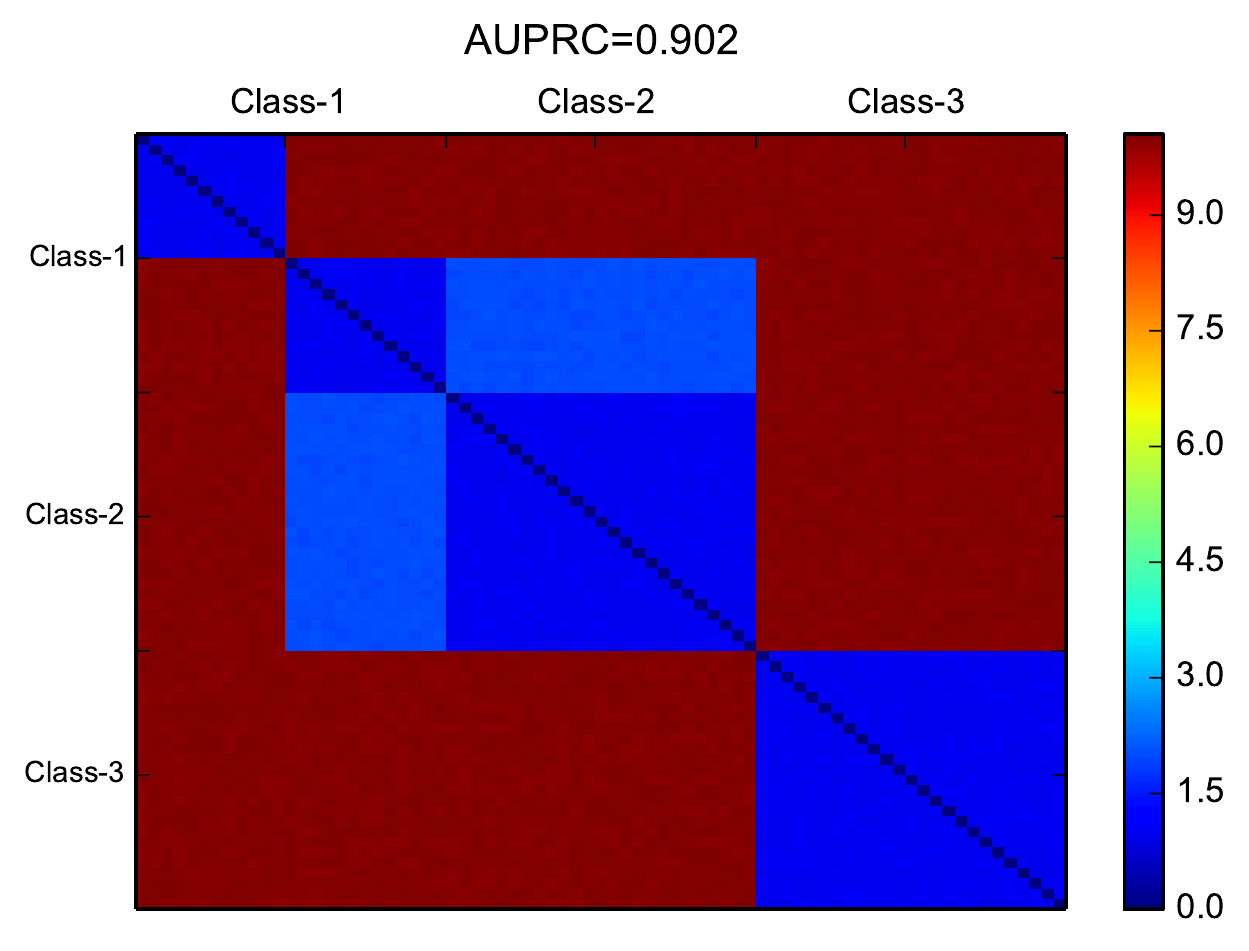}}
\subfloat[$d_3$]{\includegraphics[width=0.32\linewidth]{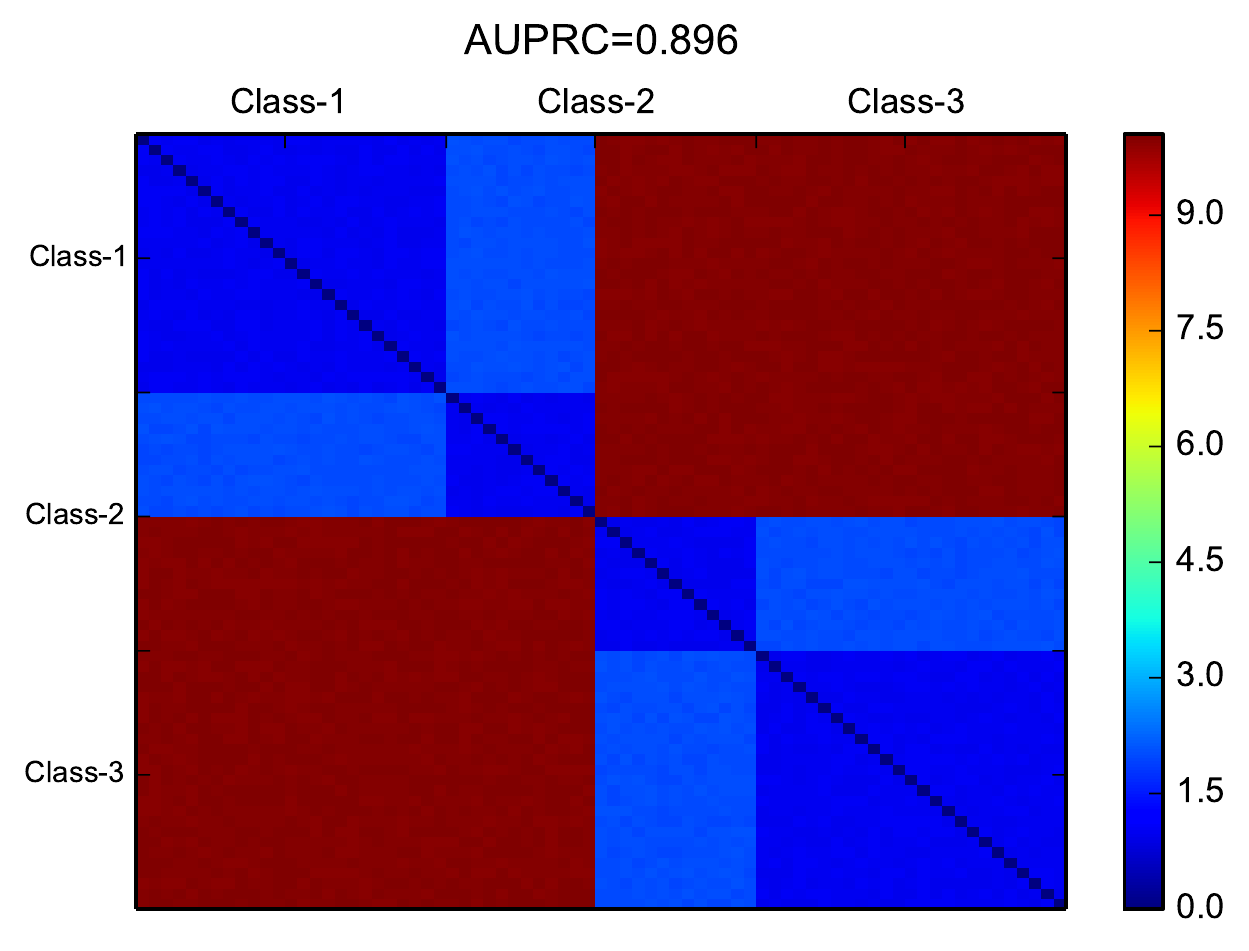}}
\caption{ Heat maps of three measures for in an example of 3 equally sized classes. a) Metric $d_1$ shows clear separation between the 3 classes. b) $d_2$ shows 3 classes with half of Class-1 positioned closer to Class-2. c) $d_3$ identifies 2 rather than 3 classes. Note that $d_1$ has lower AUPRC than $d_2$ and $d_3$ despite being best at identifying the 3 classes whereas $\overline{P}$ values for the metrics are $d_1$=1.0, $d_2$=0.887 and $d_3$=0.869. }
\label{auprcf}
\end{figure*}

Figure \ref{auprcf} gives three examples of metrics for a problem that has three classes: a) shows a metric $d_1$ (AUPRC=0.847) that clearly separates the 3-classes which, however, has a lower AUPRC than the metrics given in b) (AUPRC=0.902) which confuses half of Class-1 with Class-2 and c) (PRC=0.896) which shows 2 rather than 3 classes. The colour scale in the figure represents the magnitude of a comparison between a pair of individuals according to the corresponding metric.

Some of the problems of AUPRC are the following. First, AUPRC is based on a classifier that identifies pairs of similar networks and hence is only indirectly related to the problem of separating classes. Moreover, the classifier uses a single global threshold $\epsilon$ for all networks and classes, and hence implicitly assumes that all classes are separated on the same scale. The AUPRC further lacks a clear statistical interpretation, which complicates its use especially when one has multiple classes and when precision recall curves of different measures intersect.

Despite its problems we give AUPRC values for all measures we considered in the main text in Table \ref{auprct} for the sake of completeness. Note that $NetEmd$ measures achieve the highest AUPRC on all data sets.

\begin{table*}[!htb]
\begin{tabular}[c]{lccccc}
&$RG_1$&$RG_2$&$RG_3$&RWN&Onnela et al.\\
\midrule
$NetEmd_{G3}$&0.917$\pm$ 0.039&0.869&0.702&\bf0.800\bf &0.756\\

$NetEmd_{G4}$&0.959$\pm$ 0.030&0.930&0.759&0.774&\bf0.786\bf\\
$NetEmd_{G5}$&\bf0.981$\pm$ 0.018\bf&0.957&0.766&0.722& 0.757\\
\midrule

$NetEmd_{S}$&0.967$\pm$0.015&\bf0.958\bf&\bf0.833\bf&0.702&0.672\\
$NetEmd_{E4}$&0.966$\pm$0.030&0.945&0.801&0.777&0.739\\
$NetEmd_{DD}$&0.756$\pm$0.044&0.708&0.516&0.655&0.612\\
\midrule
$Netdis_{ER}$&0.867 $\pm$0.044 &0.579&0.396&0.607&0.621\\
$Netdis_{SF}$&0.852$\pm$0.028&0.657&0.437&0.522&0.592\\
\midrule
$GCD11$ &0.888$\pm$0.084&0.709&0.478&0.713&0.693\\
$GCD73$&0.966$\pm$0.052&0.858&0.571&0.736&0.743\\
\midrule
$GGDA$&0.815$\pm$0.176&0.740&0.481&0.500&0.625\\
\bottomrule
\end{tabular}
\caption{AUPRC scores for measures and data sets considered in the main text. $NetEmd$ measures have the highest AUPRC score (given in bold) on all data sets.For $RG_1$ we calculated the value of the AUPRC score for each of the 16 sub-data sets. The table shows the average and standard deviation of the AUPRC values obtained over these 16 sub-data sets. }
\label{auprct}
\end{table*}
\bibliography{sample}
\end{document}